\documentclass{article} 
\usepackage{iclr2027_conference,times}

\usepackage{amsmath,amsfonts,bm}

\def\eqref#1{equation~\ref{#1}}

\def\1{\bm{1}}

\DeclareMathAlphabet{\mathsfit}{\encodingdefault}{\sfdefault}{m}{sl}
\SetMathAlphabet{\mathsfit}{bold}{\encodingdefault}{\sfdefault}{bx}{n}

\usepackage{hyperref}
\usepackage{url}
\usepackage{hyperref}       
\usepackage{url}            
\usepackage{booktabs}       
\usepackage{amsfonts}       
\usepackage{nicefrac}       
\usepackage{microtype}      
\usepackage{lipsum}
\usepackage{fancyhdr}       
\usepackage{graphicx}       
\usepackage{natbib}

\usepackage{xcolor}
\usepackage{tcolorbox}
\tcbuselibrary{breakable}
\tcbuselibrary{breakable}
\usepackage{mathtools}
\usepackage{hyperref} 
\usepackage{cleveref}
\usepackage{algorithm}
\usepackage[noend]{algorithmic}
\usepackage{graphicx}
\usepackage{textcomp}
\usepackage{multirow}
\usepackage{xspace}
\usepackage{listings}
\usepackage{caption}
\usepackage{amsthm}
\usepackage{float}
\usepackage{soul}
\usepackage{upquote}
\usepackage{listings}
\usepackage{enumitem}
\usepackage{extarrows}
\usepackage{stmaryrd}
\usepackage{bbm}
\usepackage{subcaption}
\usepackage{makecell}
\usepackage{array}
\usepackage{dsfont}
\usepackage{colortbl}
\usepackage{tikz}
\usepackage{wrapfig}
\usepackage{amssymb}
\usepackage{diagbox}
\usepackage{bm}
\usepackage{fvextra}
\usepackage{thmtools}

\theoremstyle{plain}

\theoremstyle{definition}

\crefname{assumption}{assumption}{assumptions}
\Crefname{assumption}{Assumption}{Assumptions}

\newcommand{\C}{\mathcal{C}}
\newcommand{\D}{\mathcal{D}}

\newcommand{\K}{\mathcal{K}}

\newcommand{\tsf}[1]{\textsf{#1}}

\newcommand{\sys}{\tsf{LGP}\xspace}

\usepackage{xcolor}
\definecolor{gainGreen}{RGB}{34,139,34}

\newcommand{\lgpgain}[2]{%
  \textbf{#1}\textsubscript{\textcolor{gainGreen}{\scriptsize +#2\%}}%
}

\definecolor{exactGray}{gray}{0.35}

\definecolor{lgpcolor}{RGB}{0,100,70}
\newcommand{\lgpadd}[1]{\textcolor{lgpcolor}{#1}}

\usepackage{placeins}

\title{Better Nearest Neighbor Graph Indices via (Efficient) LLM-Guided Pruning}

\author{Fangzhou Wu$^{1}$, Haike Xu$^2$, Sandeep Silwal$^1$ \\
$^{1}$University of Wisconsin--Madison, $^{2}$MIT \\
\texttt{fwu89@wisc.edu, haikexu@mit.edu, silwal@cs.wisc.edu} \\
}

\iclrfinalcopy 
\begin{document}

\maketitle

\lhead{Preprint} 

\begin{abstract}

Graph-based approximate nearest neighbor search (ANNS) is widely used for large-scale semantic search.
Its indices are constructed primarily based on \emph{geometric relationships} among embeddings of an input dataset (e.g., documents or images), rather than explicitly optimizing for \emph{semantic relevance}.
However, when using these indices for downstream query retrieval, performance is evaluated based on the semantic relevance of the retrieved results to the query.
This creates a fundamental ``geometry--semantic'' mismatch between how the indices are \emph{constructed} and how their retrieval results are \emph{evaluated}.
While existing LLM-based reranking methods can partially mitigate this mismatch at query time, they leave this underlying structural problem in the graph unaddressed. 
We therefore propose \emph{\textbf{L}LM-Guided \textbf{G}raph \textbf{P}runing} (\sys), a general framework that addresses this mismatch directly by leveraging LLM reasoning to refine an existing ANN graph index itself.
\sys identifies ``low-value" neighbors of nodes and replaces them with LLM-selected alternatives that provide useful semantic information while retaining desired geometric structures of the original graph, including sparsity and efficient navigability.
Experiments on representative semantic retrieval benchmarks show that \sys consistently improves end-to-end retrieval performance over both vanilla greedy graph search and LLM-based reranking across widely used graph-based ANN indices such as DiskANN and HNSW.
Our code is available at \url{https://github.com/fzwark/LGP}
\end{abstract}

\section{Introduction}
Graph-based Approximate Nearest Neighbor Search (ANNS) is a popular and foundational component of modern retrieval systems~\citep{11202651}, including retrieval-augmented search~\citep{ARSLAN20243781}, and is an active area of research in the machine learning, algorithms, and database communities.\footnote{See, e.g., the recent VecDB Workshop at VLDB26 (\url{https://vecdb-ws.github.io/vldb2026/}).}
At a high level, this family of algorithms constructs a sparse, navigable graph (i.e., one that enables greedy search to efficiently reach points close to a query) over an input dataset (e.g., documents or images) during preprocessing, based on the geometric proximity of their embeddings~\cite{10.1109/TPAMI.2018.2889473,NEURIPS2019_09853c7f}.  
When a query arrives, the prevailing paradigm is to perform a (highly efficient) greedy search over this graph based on embedding similarity, followed by a reranking step over the retrieved candidates. 
This second step often utilizes a Large Language Model (LLM) or a more accurate similarity computation and is much more computationally expensive than the greedy search~\citep{sun-etal-2023-chatgpt,pradeep2023rankvicunazeroshotlistwisedocument,qin-etal-2024-large,weller2025rank,shao2025reasonir}. 
However, this reranking step is considered crucial because embeddings alone often fail to fully capture the complex semantic relationships between queries and the input dataset, and thus, a more ``powerful'' secondary semantic evaluator is necessary for accurate retrieval~\citep{chan2024rqrag,inf-x-retriever-2025}. 

From this discussion, it is evident that there is a fundamental mismatch between \emph{how graph indices are built and how they are ultimately deployed}. 
They are built using embedding-space geometry due to efficiency constraints, but actual query search requires refining the greedy search using LLMs or expensive similarity measures for the sake of accuracy. 
This ``geometry--semantic'' mismatch 
is especially problematic under a limited search budget, where graph search can visit only a small fraction of the input dataset~\citep{yoon2025acurankuncertaintyawareadaptivecomputation}.
Within this budget, the graph search may favor geometrically close but semantically less relevant documents while missing relevant ones.
Even though ``post-hoc'' methods such as LLM-based reranking can partially mitigate this problem, their effectiveness remains constrained by the recall of relevant documents in the initial candidate set.
If a relevant document is never reached initially, even a powerful LLM reranker cannot recover it afterward.
Although increasing the search budget can reduce this ``candidate bottleneck", doing so incurs prohibitive query-time computation.

Another possibility is to construct the graph directly according to semantic relationships among documents.
However, evaluating these relationships across a large dataset using expensive measures, such as LLM-based reasoning, would again incur substantial computational costs.
More importantly, a graph created based on purely semantic relationships may no longer be compatible with the efficient geometric computations used by greedy search at query time.
Then navigating such a graph using greedy search may therefore reduce retrieval accuracy, whereas using stronger semantic measures to guide navigation would substantially increase query-time latency and cost. \textbf{Our paper focuses on addressing this mismatch between geometry-driven graph construction and downstream semantic retrieval under a constrained search budget}.

\noindent\textbf{Our Results.}
To understand how pronounced this mismatch is in practice and whether geometry and semantics can be better balanced in a single graph, we first conduct a controlled and idealized empirical analysis. 
Our analysis shows that graph construction based on embeddings alone leaves substantial semantic search potential untapped: changing only a single outgoing edge to a local alternative (from a two-hop neighborhood) can substantially improve ground-truth recall under the same search budget, while preserving useful geometric properties.
These findings motivate performing targeted, ``local'' edge updates on top of an existing graph index to better reconfigure its edge connections with semantic relationships, while still preserving efficient navigation.

Based on these insights, we propose \emph{\textbf{L}LM-\textbf{G}uided Graph \textbf{P}runing} (\sys), a general framework that moves beyond only using reranking at query time and uses LLM-guided graph refinement \emph{during the index construction itself}. 
\sys operates on top of existing embedding-based graph indices and makes local edge changes. At a high level, \sys constructs a bounded candidate pool from local two-hop neighborhoods, identifies structurally redundant neighbors that provide little additional reachability capabilities, and judiciously uses LLM reasoning to select candidates that add distinct semantic value beyond the current neighborhood. These new candidates replace the identified ``low-value'' neighbors.
Crucially, \sys is designed to preserve the geometric structure that supports efficient graph search (i.e., navigability) and leaves the query-time greedy search procedure unchanged, while making local neighborhoods more semantically informative without sacrificing sparsity or introducing additional LLM inference at query time. 
Furthermore, since it modifies only the graph index, it can be seamlessly integrated into existing retrieve-then-rerank pipelines.

To show its versatility, we instantiate \sys on two widely used graph-based ANN indices, DiskANN~\citep{NEURIPS2019_09853c7f} and HNSW~\citep{10.1109/TPAMI.2018.2889473}, although the framework can be applied more broadly to other graph-based ANN methods.
Extensive experiments demonstrate that \sys addresses limitations of prior embedding-based graph constructions and improves end-to-end retrieval performance.
Specifically, relative to the vanilla baseline, \sys improves average greedy-search NDCG@10 by \textbf{24.4\%} on the textual BRIGHT and \textbf{11.7\%} on the multimodal M-BEIR, with the improvements reaching \textbf{26.0\%} and \textbf{13.2\%} after reranking, respectively.
Furthermore, \sys substantially improves search efficiency, achieving comparable retrieval quality with up to \textbf{50\% fewer distance computations}.
To summarize, our main contributions are:
\begin{enumerate}[leftmargin=*, noitemsep]
    \item We identify and formulate a fundamental \emph{geometry--semantic mismatch} in graph-based ANN retrieval, where graph indices are constructed based on embedding-space geometry, while downstream retrieval is ultimately evaluated by semantic relevance. 
    \item Through controlled one-edge replacements, we empirically show that geometrically comparable edge choices give substantially different ground-truth recall during search, revealing the potential for better trade-offs between geometric navigation and semantic retrieval in a single graph.
    \item Guided by these findings, we propose \sys, an offline graph-refinement framework that selectively replaces structurally low-value neighbors with LLM-selected alternatives that provide more useful semantic connections, while leaving the query-time search procedure unchanged.
    \item We instantiate \sys on two widely used graph-based ANN indices, DiskANN and HNSW, and demonstrate consistent retrieval improvements across semantic retrieval benchmarks while preserving graph sparsity and achieving comparable retrieval quality with up to {50\% fewer distance computations}.
\end{enumerate}

\section{Preliminaries}
Graph-based approximate nearest neighbor search (ANNS) methods build sparse navigable graphs over dense vector representations to support efficient retrieval~\citep{NEURIPS2019_09853c7f,10.1109/TPAMI.2018.2889473}. 
Here, navigability refers to the ability of greedy graph search to efficiently reach nodes that are close to a query in embedding space.
Given a corpus $\D=\{d_1,\dots,d_{|\D|}\}$, each document $d_i \in \D$ is encoded by a bi-encoder into an embedding vector $\bm e(d_i)$. 
A graph-based ANN index is represented as a directed graph $G=(V,E)$, where each node $v_i\in V$ corresponds to a document $d_i$, and $(v_i,v_j)\in E$ denotes a directed edge from $v_i$ to $v_j$.
We use $N(i)=\{v_j:(v_i,v_j)\in E\}$ to denote the outgoing neighborhood of $v_i$ with degree bound $R$.
We use $D(\cdot,\cdot)$ to denote the embedding-space distance (e.g., cosine distance) between two nodes or between a query and a node.


Although graph-based ANN indices differ in construction details, most follow a common pattern.
For each node $v_i$, the index first retrieves a candidate set $U_i$ through graph traversal or local search, then applies an index-specific neighbor-selection rule to form the outgoing neighborhood $N(i)$ under degree limit $R$.
For example, DiskANN uses geometric pruning on a flat graph, while HNSW combines hierarchical search with heuristic neighbor selection.\footnote{Detailed algorithms for DiskANN and HNSW are provided in~\Cref{app:original}.}
In both cases, construction is based on embedding-space geometry rather than explicit semantic relationships among documents.

At query time, a query $q$ is encoded as $\bm e(q)$ and greedy search is performed over $G$.
Starting from an entry node, the search repeatedly expands promising frontier nodes according to their embedding-space distance to $q$ and evaluates their outgoing neighbors.
We count each evaluation of the embedding-space distance between $q$ and a previously unseen node as one query--node distance computation.
Under a finite computation budget, the search explores only part of the graph, producing candidates that are either returned directly or passed to a downstream reranker.


\noindent\textbf{Other Related Work on LLM-Augmented and Query-Time Retrieval Algorithms.}
A separate line of work improves retrieval quality by adding stronger semantic reasoning at inference time, including query rewriting, reasoning-aware retrievers, and reranking methods~\citep{ma-etal-2023-query,chan2024rqrag,shao2025reasonir,zhang-etal-2025-rearank,liu2025reasonrankempoweringpassageranking}. 
Other recent methods modify graph search itself: bi-metric search uses a cheap proxy metric for indexing while adapting query-time search toward a more expensive target metric~\citep{xu2024bimetricframeworkfastsimilarity}, and RGS alters graph traversal to decide which candidates should be evaluated by a stronger reranker under a limited budget~\citep{xu2026beyond}. 
These approaches are complementary to ours: they primarily improve how a fixed graph is queried online, whereas our method improves the graph structure itself offline. 
Thus, our refined graph remains compatible with standard greedy search and a broad class of query-time retrieval algorithms that incorporate stronger semantic reasoning.
See extended related work in~\Cref{app:related}.

\section{Motivation: The Geometry--Semantic Mismatch}
\label{sec:analysis}

\begin{figure}[!t]
  \centering
  \includegraphics[width=\linewidth]{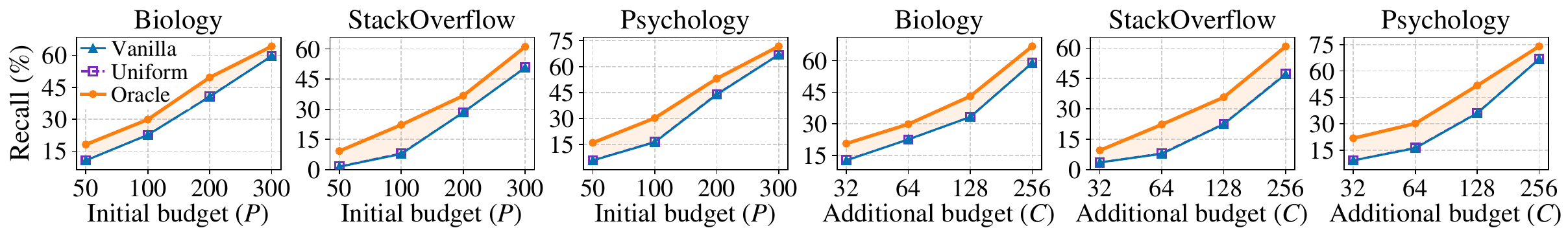}
  \caption{
Alternate one-edge replacements reveal a substantial geometry--semantic mismatch and potential for improvement.
Across both additional budgets $C$ and initial budgets $P$, \emph{Oracle} consistently achieves higher ground-truth recall than \emph{Vanilla}, while \emph{Uniform} is close to \emph{Vanilla}.
}
  \label{fig:geometry-semantic-headroom}
\vspace{-5pt}
\end{figure}

Embedding-based graph indices connect each document to a small set of geometrically close neighbors, which may not align with its underlying semantic relationships.
This creates a mismatch because geometrically preferred edges are not necessarily the most useful for semantic retrieval.
It is therefore important to empirically investigate and quantify how pronounced this mismatch is in practice.
Furthermore, sparse ANN indices generally allow multiple edge choices that are similar under the geometric constraints used to build the index, so we naturally ask: can some of these alternative edge choices be a better fit with semantic relationships, and if so, to what extent?
To this end, we design a controlled setup where we modify only a single edge of the vanilla index using alternatives from a bounded candidate set under the same geometric constraints.
This tests whether vanilla geometric construction leaves room for improving semantic retrieval without violating existing geometric constraints, and whether geometry and semantics can be balanced within a single graph.

\noindent\textbf{One-Edge Replacement.}
Specifically, for each query $q$ and initial search budget $P$, we run greedy search on the Vanilla DiskANN graph until at least $P$ unique embedding-space distance computations have been performed, and save the search immediately before the next frontier node $v_i$ is expanded.
Let $N(i)$ denote the outgoing neighbors of $v_i$, and let $N_2(i)$ denote its two-hop non-neighbors:
$
  N_2(i) = \left(\bigcup_{v_j\in N(i)}N(j)\right)\setminus\left(N(i)\cup\{v_i\}\right).
  \label{eq:two-hop-candidates}
$
Starting from this same point, for every $z\in N(i)$ and $c\in N_2(i)$, we construct a one-edge variant of the graph by replacing $(v_i,z)$ with $(v_i,c)$ and continue the greedy search for an additional budget of $C$ query--document distance computations.
By varying both $P$ and $C$, we examine whether a single local edge change can improve relevant-document discovery across different stages of the search and different search budgets.

\noindent\textbf{Measuring the Geometry--Semantic Gap.}
For each query $q$ and each pair of initial and additional search budgets $(P,C)$, we evaluate the vanilla graph and every one-edge variant by the fraction of ground-truth relevant documents, $\operatorname{GT}(q)$, reached after the additional $C$ distance computations.
Specifically, for query $q$ and graph $G$ 
, let $\D^{\mathrm{seen}}_{G}(q,P,C)$ denote the documents evaluated during the initial $P$ distance computations together with those evaluated during the additional $C$ computations.
We measure ground-truth recall as
\begin{equation}
        \frac{\left|\D^{\mathrm{seen}}_{G}(q,P,C)\cap \operatorname{GT}(q)\right|}{\left|\operatorname{GT}(q)\right|}.
\end{equation}
To preserve the geometric constraints of the original graph, we keep only one-edge variants whose recall of the exact top-10 embedding-space neighbors of $q$ is no worse than that of the vanilla graph under the same $(P,C)$ budgets.
For each query and budget setting, we define the variant with the highest ground-truth recall as \emph{Oracle} and compare it with \emph{Vanilla} to quantify the potential improvement under the same geometric constraints.
As a control, we report the average recall across all such one-edge variants as the \emph{Uniform} baseline.

\noindent\textbf{Geometry Leaves Semantic Retrieval Potential Untapped.}
We conduct this analysis on three BRIGHT datasets, Biology, StackOverflow, and Psychology, using DiskANN as the underlying graph index~\citep{su2025bright}.
As shown in Figure~\ref{fig:geometry-semantic-headroom}, \emph{Oracle} consistently outperforms \emph{Vanilla} in ground-truth recall across varying $C$ at fixed $P=100$ and varying $P$ at fixed $C=64$.
This consistent gap confirms that geometrically comparable edge choices can differ substantially in their semantic retrieval performance, directly demonstrating a substantial geometry--semantic mismatch in practice.
More importantly, these results indicate that graph construction based on embeddings alone leaves substantial potential for improving semantic relevance while satisfying the same geometric constraints.
Furthermore, the near-zero improvement of \emph{Uniform} shows that arbitrary one-edge replacements do not perform well, and that realizing this potential requires selecting semantically useful alternatives.
These findings suggest that geometry and semantics can be combined and ``harmonized'' within a single graph and motivate our method, \sys, which selectively prunes local edges on top of the existing geometric index to incorporate more semantically useful connections.

\section{Our Framework: LLM-Guided Graph Pruning}\label{sec:method}

\begin{algorithm}[t]
\small
\caption{LLM-Guided Graph Pruning (\sys)}
\label{alg:llm_pruning_general}
\begin{algorithmic}[1]
\STATE \textbf{Input} ANN graph $G$, current node $v_i$, neighborhood $N(i)$, and index-specific candidate set $U_i$
\STATE \textbf{Parameters} Degree cap $R$, shortlist budget $M$, and edit budget $B$
\STATE {\ttfamily \textcolor{red}{\(\triangleright\) {/* Form a local replacement set */}}}
\STATE $U_i \leftarrow U_i$ together with directed two-hop non-neighbors of $v_i$
\STATE {\ttfamily \textcolor{red}{\(\triangleright\) {/* Shortlist local replacement candidates */}}}
\STATE $U_i \leftarrow$ shortlist up to $M$ candidates based on additional two-hop reachability and redundancy with $N(i)$
\STATE {\ttfamily \textcolor{red}{\(\triangleright\) {/* Select semantically useful candidates with an LLM */}}}
\STATE $\mathcal{A}_i \leftarrow \textsc{LLMSelect}(v_i, N(i), U_i, B)$
\STATE {\ttfamily \textcolor{red}{\(\triangleright\) {/* Local edge pruning and replacement */}}}
\FOR{each $v_c \in \mathcal{A}_i$}
    \STATE Identify an existing neighbor $v_w$ with little additional reachability and high redundancy
    \STATE Update $N(i)$ by adding $v_c$ and, when necessary, removing $v_w$ to satisfy the degree constraint
\ENDFOR
\end{algorithmic}
\end{algorithm}




To address the geometry--semantic mismatch while preserving efficient graph navigation, we introduce \emph{\textbf{L}LM-Guided \textbf{G}raph \textbf{P}runing} (\sys), a general framework for refining an existing ANN index by making its local neighborhoods better aligned with semantic relationships (\Cref{alg:llm_pruning_general}).
Rather than replacing the original indexing algorithm, \sys operates on top of the existing graph index (e.g., DiskANN and HNSW), identifies neighbors that provide little additional reachability beyond the rest of the local neighborhood, and replaces them with LLM-selected candidates that can help the search reach more relevant documents without violating the geometric constraints of the original index.
For each node in the index, our algorithm proceeds in three main steps.
First, it constructs a small set of local replacement candidates using directed two-hop non-neighbors.
It then uses efficient to compute ``structural" signals, including additional two-hop reachability and neighborhood redundancy, to shortlist promising candidates.
From this shortlist, LLM reasoning is used to select semantically useful alternatives.
Finally, the selected candidates are added to the neighborhood by replacing existing neighbors with little additional reachability and high redundancy.
A detailed version of the algorithm is provided in~\Cref{alg:llm_pruning_detailed} in the appendix.


\noindent\textbf{Efficient Local Candidate Set Construction.}
Given an ANN index $G$, for each node $v_i$, \sys considers only a small set of local replacement candidates rather than all nodes in the graph.
Since most graph-based ANN methods already produce a candidate set of geometrically close neighbors when selecting or updating the outgoing neighborhood $N(i)$ of $v_i$, \sys reuses this set, denoted by $U_i$, and augments it with directed two-hop non-neighbors, i.e., nodes reachable through a current neighbor but not already directly connected to $v_i$.
This provides a broader set of local alternatives from which the LLM can identify semantically useful replacements, while keeping candidate construction computationally inexpensive.
To further control the cost and context length of LLM selection, \sys prunes the augmented set to at most $M$ candidates using only graph connectivity and embedding distances, without invoking the LLM.
The pruning favors candidates that provide greater additional two-hop reachability and are less redundant with the current neighborhood (lines 7--15 in~\Cref{alg:llm_pruning_detailed}).

\noindent\textbf{LLM-Guided Semantic Selection.}
From the shortlisted set $U_i$, \sys uses an LLM to identify replacement candidates that add useful semantic information beyond what is already represented in the current neighborhood.
The LLM is given the document represented by $v_i$, a small set of its current neighbors as context, and the documents in the shortlisted set $U_i$.
It selects up to $B$ candidates that extend rather than repeat the information already available through the current neighborhood, for example, by providing missing supporting context or connecting $v_i$ to a related concept that is not well represented by its existing neighbors.
Here, $B$ is a small per-node edit budget that limits the number of edge changes (e.g., $B=2$).
Prompts for \textsc{LLMSelect} are provided in~\Cref{app:prompts}.

\begin{wrapfigure}[10]{R}{0.58\textwidth}
    \vspace{-1.2em}
    \centering
    \includegraphics[width=\linewidth]{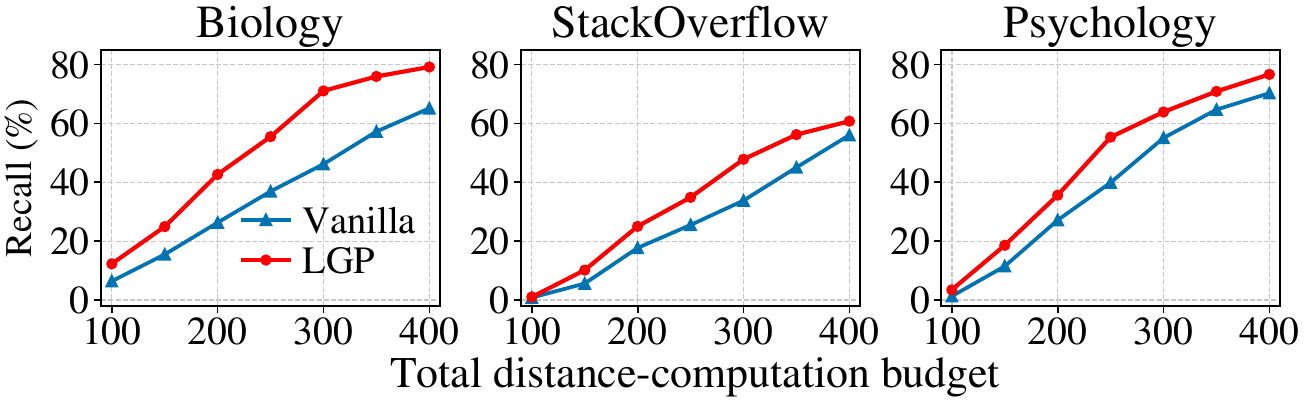}
    \caption{
    \sys-refined DiskANN consistently improves recall over Vanilla across distance-computation budgets.
    }
    \label{fig:vanilla-lgp-prefix}
\end{wrapfigure}

\noindent\textbf{Local Edge Replacement.}
After LLM-guided selection, \sys applies a lightweight structural check to keep candidates that provide sufficient additional reachability.
Such a candidate is added directly when the neighborhood is below the degree limit.
Otherwise, \sys replaces an existing neighbor that provides little additional two-hop reachability beyond the other neighbors and is highly redundant with the current neighborhood.
This allows \sys to introduce more semantically useful connections while preserving the sparsity and local geometric structure of the original index.

\noindent\textbf{Instantiation for Representative ANN Indices.}
\sys is general and can be instantiated on different graph-based ANN indices, including DiskANN and HNSW.
In DiskANN, \sys is applied immediately after RobustPrune during the second construction pass, taking the nodes visited by GreedySearch as $U_i$.
In HNSW, \sys refines only the base layer of the constructed index, taking the base-layer nodes visited by hierarchical search as $U_i$.
Both instantiations follow the shared pruning procedure in~\Cref{alg:llm_pruning_general}, with detailed algorithms provided in~\Cref{app:ins}.


\noindent\textbf{\sys Realizes the Semantic Improvement Potential.}
To examine whether the improvement potential identified by the controlled one-edge analysis in~\Cref{sec:analysis} can be realized by \sys, we compare the fixed \sys-refined DiskANN index directly against the original vanilla index across total distance-computation budgets.
As shown in~\Cref{fig:vanilla-lgp-prefix}, \sys consistently achieves higher ground-truth recall than \emph{Vanilla} across all tested budgets and datasets.
These results show that the potential revealed by the one-edge analysis translates into consistent retrieval gains with \sys, demonstrating that semantic relevance and geometric constraints can be better balanced within a single graph.

\section{Empirical Results}\label{sec:experiments}

\begin{table*}[t]
\centering
\small
\setlength{\tabcolsep}{3.5pt}
\renewcommand{\arraystretch}{1.04}
\caption{
BRIGHT main results on DiskANN at search width 20 using Diver embeddings.
Each entry reports Vanilla\,/\,\sys.
\textcolor{gainGreen}{Green} subscripts denote relative improvements over Vanilla.
}
\label{tab:bright-main}
{
\begin{tabular}{lcccc}
\toprule
& \multicolumn{2}{c}{Greedy search}
& \multicolumn{2}{c}{Rerank} \\
\cmidrule(lr){2-3}\cmidrule(lr){4-5}
Dataset
& NDCG@10
& Recall@10
& NDCG@10
& Recall@10 \\
\midrule

Biology
& 24.69\,/\,\lgpgain{37.06}{50.1}
& 28.66\,/\,\lgpgain{44.63}{55.7}
& 28.30\,/\,\lgpgain{41.57}{46.9}
& 30.71\,/\,\lgpgain{46.19}{50.4}
\\

Earth Science
& 20.60\,/\,\lgpgain{27.92}{35.5}
& 22.70\,/\,\lgpgain{30.59}{34.8}
& 23.29\,/\,\lgpgain{30.06}{29.1}
& 25.51\,/\,\lgpgain{33.03}{29.5}
\\

Economics
& 19.56\,/\,\lgpgain{22.35}{14.3}
& 20.77\,/\,\lgpgain{25.61}{23.3}
& 19.98\,/\,\lgpgain{25.42}{27.2}
& 21.92\,/\,\lgpgain{28.84}{31.6}
\\

Psychology
& 28.31\,/\,\lgpgain{34.76}{22.8}
& 34.51\,/\,\lgpgain{41.61}{20.6}
& 32.78\,/\,\lgpgain{41.46}{26.5}
& 34.40\,/\,\lgpgain{44.10}{28.2}
\\

Robotics
& 19.48\,/\,\lgpgain{21.68}{11.3}
& 24.83\,/\,\lgpgain{27.12}{9.2}
& 25.67\,/\,\lgpgain{28.63}{11.5}
& 28.61\,/\,\lgpgain{31.16}{8.9}
\\

StackOverflow
& 10.82\,/\,\lgpgain{17.36}{60.4}
& 13.87\,/\,\lgpgain{21.37}{54.1}
& 11.96\,/\,\lgpgain{19.61}{64.0}
& 14.66\,/\,\lgpgain{23.73}{61.9}
\\

Sustainable Living
& 18.09\,/\,\lgpgain{24.33}{34.5}
& 22.12\,/\,\lgpgain{29.76}{34.5}
& 20.20\,/\,\lgpgain{27.43}{35.8}
& 22.04\,/\,\lgpgain{30.82}{39.8}
\\

Pony
& 17.08\,/\,\lgpgain{17.80}{4.2}
& 9.06\,/\,\lgpgain{9.40}{3.8}
& 19.54\,/\,\lgpgain{20.07}{2.7}
& \textbf{9.73}\,/\,9.00
\\

LeetCode
& 3.20\,/\,\lgpgain{6.20}{93.8}
& 4.54\,/\,\lgpgain{9.51}{109.5}
& 2.97\,/\,\lgpgain{6.37}{114.5}
& 4.27\,/\,\lgpgain{10.07}{135.8}
\\

AoPS
& 3.13\,/\,\lgpgain{3.36}{7.3}
& 4.58\,/\,\lgpgain{5.03}{9.8}
& 3.12\,/\,\lgpgain{3.23}{3.5}
& 4.69\,/\,\lgpgain{4.79}{2.1}
\\

TheoremQA-Theorems
& 34.70\,/\,\lgpgain{36.35}{4.8}
& 46.26\,/\,\lgpgain{47.57}{2.8}
& 33.57\,/\,\lgpgain{36.44}{8.5}
& 45.60\,/\,\lgpgain{48.89}{7.2}
\\

TheoremQA-Questions
& 5.73\,/\,\lgpgain{6.45}{12.6}
& 8.24\,/\,\lgpgain{8.36}{1.5}
& 5.63\,/\,\lgpgain{5.68}{0.9}
& 9.24\,/\,\lgpgain{9.49}{2.7}
\\

\midrule

Average
& 17.12\,/\,\lgpgain{21.30}{24.4}
& 20.01\,/\,\lgpgain{25.05}{25.2}
& 18.92\,/\,\lgpgain{23.83}{26.0}
& 20.95\,/\,\lgpgain{26.68}{27.4}
\\

\bottomrule
\end{tabular}
}
\vspace{-13pt}
\end{table*}

We comprehensively evaluate \sys on textual and multimodal benchmarks using DiskANN and HNSW.
Across settings, \sys consistently improves greedy-search retrieval, and these gains persist after downstream reranking.
The improvements are most pronounced at narrow search widths.

\noindent\textbf{Benchmarks and Models.}
We use BRIGHT~\citep{su2025bright} for reasoning-intensive text retrieval and M-BEIR~\citep{10.1007/978-3-031-73021-4_23} for multimodal retrieval.
BRIGHT contains 12 datasets spanning diverse domains, while M-BEIR contains 16 tasks covering text, image, and image--text retrieval.
We evaluate all BRIGHT queries and, following~\citet{xu2026beyond}, uniformly sample 100 queries per M-BEIR task using the same samples across all experiments.
For BRIGHT, we use Diver-Retriever-4B-1020~\citep{sun2026divermultistageapproachreasoningintensive} and Qwen3-Embedding-4B as embedding models.
For M-BEIR, following~\citet{xu2026beyond}, we use CLIP (ViT-B/32)~\citep{pmlr-v139-radford21a} to encode all 16 task subsets.
For LLM-guided semantic selection and downstream reranking, we use Qwen3-32B~\citep{qwen3technicalreport} on BRIGHT and Qwen3-VL-30B-A3B-Instruct~\citep{bai2025qwen3vltechnicalreport} on M-BEIR, both in non-reasoning mode.
Detailed benchmark and model configurations are provided in~\Cref{tab:bright-populations,tab:mbeir-populations,tab:experiment-models}.

\noindent\textbf{Index Configurations.}
Our primary ANN index is DiskANN, with a maximum out-degree of 16, a construction search width
of 125, and a pruning parameter of 1.2.
We additionally evaluate HNSW with connectivity 16 and construction search width 200.
We use an edit budget of $B=2$ for both benchmarks, with a shortlist size of $M=24$ on BRIGHT and $M=12$ on M-BEIR.

\noindent\textbf{Retrieval Configurations and Metrics.}
At query time, we evaluate greedy graph search and downstream reranking.
We use \emph{search width}, the standard query-time parameter for controlling search effort in graph-based ANN indices, corresponding to $L_{\mathrm{search}}$ in DiskANN and $ef_{\mathrm{search}}$ in HNSW.
With graph degree bounded by $R$, the query-time search cost grows roughly with $\text{search width}\times R$ in terms of distance computations.
For DiskANN, we vary $L_{\mathrm{search}}\in\{10,20,50,100,200\}$; for HNSW, we vary $ef_{\mathrm{search}}\in\{10,12,14,16,18,20,50\}$, using 20 by default.
Greedy search returns candidate documents ordered by embedding distance.
We either evaluate the top $k$ directly or rerank the same candidate set before evaluating the resulting top $k$.
We report NDCG@$k$ and document-level Recall@$k$ for $k\in\{5,10\}$ on a 0--100 scale.
Recall@$k$ is the fraction of all ground-truth documents that appear in the top $k$, averaged over queries.
When a query has more than $k$ ground-truth documents, the maximum possible Recall@$k$ is below 100.
Full details are provided in~\Cref{app:exp_setup}.

\begin{table*}[t]
\centering
\small
\setlength{\tabcolsep}{4pt}
\renewcommand{\arraystretch}{1.04}
\caption{
M-BEIR results on DiskANN at search width 20 using CLIP.
Each entry reports Vanilla/\sys.
}
\label{tab:mbeir-main}
{
\begin{tabular}{lcccc}
\toprule
& \multicolumn{2}{c}{Greedy search}
& \multicolumn{2}{c}{Rerank} \\
\cmidrule(lr){2-3}\cmidrule(lr){4-5}
Dataset (Task)
& NDCG@10
& Recall@10
& NDCG@10
& Recall@10 \\
\midrule

VisualNews (0)
& \textbf{16.41}\,/\,15.45
& \textbf{21.00}\,/\,20.00
& 15.23\,/\,\lgpgain{15.97}{4.9}
& 21.00\,/\,21.00
\\

MSCOCO (0)
& 32.45\,/\,\lgpgain{32.88}{1.3}
& 51.00\,/\,\lgpgain{52.00}{2.0}
& 45.09\,/\,\lgpgain{46.09}{2.2}
& 56.00\,/\,\lgpgain{57.00}{1.8}
\\

Fashion200K (0)
& 1.80\,/\,\lgpgain{1.82}{1.1}
& 2.63\,/\,2.63
& \textbf{4.53}\,/\,3.56
& \textbf{4.97}\,/\,3.97
\\

WebQA (1)
& 5.99\,/\,\lgpgain{6.61}{10.4}
& 7.50\,/\,\lgpgain{8.00}{6.7}
& 8.29\,/\,\lgpgain{8.93}{7.7}
& 8.00\,/\,\lgpgain{9.00}{12.5}
\\

EDIS (2)
& 6.76\,/\,\lgpgain{11.29}{67.0}
& 9.07\,/\,\lgpgain{14.74}{62.5}
& 11.02\,/\,\lgpgain{15.67}{42.2}
& 12.80\,/\,\lgpgain{18.71}{46.2}
\\

WebQA (2)
& 15.15\,/\,\lgpgain{22.97}{51.6}
& 20.50\,/\,\lgpgain{29.50}{43.9}
& 20.25\,/\,\lgpgain{29.18}{44.1}
& 23.50\,/\,\lgpgain{33.50}{42.6}
\\

VisualNews (3)
& 3.87\,/\,\lgpgain{7.04}{81.9}
& 9.00\,/\,\lgpgain{14.00}{55.6}
& 5.25\,/\,\lgpgain{10.29}{96.0}
& 11.00\,/\,\lgpgain{19.00}{72.7}
\\

MSCOCO (3)
& 42.42\,/\,\lgpgain{43.00}{1.4}
& 42.60\,/\,\lgpgain{42.80}{0.5}
& 52.63\,/\,\lgpgain{53.92}{2.5}
& 49.20\,/\,\lgpgain{49.80}{1.2}
\\

Fashion200K (3)
& \textbf{2.93}\,/\,2.86
& 6.00\,/\,6.00
& 3.97\,/\,\lgpgain{4.00}{0.8}
& 6.00\,/\,6.00
\\

NIGHTS (4)
& 28.73\,/\,28.73
& 51.00\,/\,51.00
& 41.99\,/\,\lgpgain{42.51}{1.2}
& 61.00\,/\,61.00
\\

OVEN (6)
& 0.00\,/\,0.00
& 0.00\,/\,0.00
& 0.00\,/\,0.00
& 0.00\,/\,0.00
\\

InfoSeek (6)
& 0.00\,/\,0.00
& 0.00\,/\,0.00
& 0.00\,/\,0.00
& 0.00\,/\,0.00
\\

CIRR (7)
& 6.47\,/\,\lgpgain{7.83}{21.0}
& 13.00\,/\,\lgpgain{16.00}{23.1}
& 10.99\,/\,\lgpgain{12.34}{12.3}
& 13.00\,/\,\lgpgain{15.00}{15.4}
\\

FashionIQ (7)
& 0.67\,/\,\lgpgain{1.12}{67.2}
& 2.00\,/\,\lgpgain{3.00}{50.0}
& 1.61\,/\,\lgpgain{3.66}{127.3}
& 4.00\,/\,\lgpgain{6.00}{50.0}
\\

OVEN (8)
& 0.50\,/\,\lgpgain{1.35}{170.0}
& 0.75\,/\,\lgpgain{2.08}{177.3}
& 1.16\,/\,\lgpgain{3.09}{166.4}
& 0.75\,/\,\lgpgain{2.08}{177.3}
\\

InfoSeek (8)
& 0.48\,/\,\lgpgain{0.93}{93.8}
& 1.00\,/\,\lgpgain{1.58}{58.0}
& 0.45\,/\,\lgpgain{2.53}{462.2}
& 1.00\,/\,\lgpgain{2.31}{131.0}
\\

\midrule

Average
& 10.29\,/\,\lgpgain{11.49}{11.7}
& 14.82\,/\,\lgpgain{16.46}{11.1}
& 13.90\,/\,\lgpgain{15.73}{13.2}
& 17.01\,/\,\lgpgain{19.02}{11.8}
\\

\bottomrule
\end{tabular}
}
\vspace{-5pt}
\end{table*}
\begin{table*}[t]
\centering
\small
\setlength{\tabcolsep}{1.5pt}
\renewcommand{\arraystretch}{1.04}
\caption{
Search-width sweeps on BRIGHT and M-BEIR with Diver and CLIP embeddings, respectively.
\textcolor{gainGreen}{Green} subscripts denote relative improvements over the Vanilla baseline.
\textcolor{exactGray}{Gray} values denote the average exact-$k$NN NDCG@10\,/\,Recall@10 for each benchmark.
}
\label{tab:search-width}
{
\begin{tabular}{lccccccccc}
\toprule
\multirow{3}{*}[-1ex]{
  \makecell[c]{
    Benchmark\\[-1pt]
    \footnotesize\textcolor{exactGray}{(Exact $k$NN ref.)}
  }
}
 & \multirow{3}{*}[-1ex]{\makecell[c]{Search\\width}}
 & \multicolumn{4}{c}{Greedy search}
 & \multicolumn{4}{c}{Rerank} \\
\cmidrule(lr){3-6}\cmidrule(lr){7-10}
 & & \multicolumn{2}{c}{NDCG@10}
   & \multicolumn{2}{c}{Recall@10}
   & \multicolumn{2}{c}{NDCG@10}
   & \multicolumn{2}{c}{Recall@10} \\
\cmidrule(lr){3-4}\cmidrule(lr){5-6}
\cmidrule(lr){7-8}\cmidrule(lr){9-10}
 & & Vanilla & \sys
   & Vanilla & \sys
   & Vanilla & \sys
   & Vanilla & \sys \\
\midrule
\multirow{5}{*}{
  \makecell[l]{
    BRIGHT\\[-1pt]
    \footnotesize \textcolor{exactGray}{(28.50\,/\,34.18)}
  }
}
 & 10
 & 11.90 & \lgpgain{15.74}{32.3}
 & 13.53 & \lgpgain{18.27}{35.0}
 & 13.01 & \lgpgain{17.28}{32.8}
 & 13.53 & \lgpgain{18.27}{35.0} \\
 & 20
 & 17.12 & \lgpgain{21.30}{24.4}
 & 20.01 & \lgpgain{25.05}{25.2}
 & 18.92 & \lgpgain{23.83}{26.0}
 & 20.95 & \lgpgain{26.68}{27.4} \\
 & 50
 & 23.09 & \lgpgain{25.03}{8.4}
 & 27.43 & \lgpgain{29.51}{7.6}
 & 26.31 & \lgpgain{28.47}{8.2}
 & 29.95 & \lgpgain{32.56}{8.7} \\
 & 100
 & 26.10 & \lgpgain{26.87}{3.0}
 & 31.05 & \lgpgain{31.73}{2.2}
 & 30.24 & \lgpgain{31.28}{3.4}
 & 34.94 & \lgpgain{35.61}{1.9} \\
 & 200
 & 26.94 & \lgpgain{27.42}{1.8}
 & 32.10 & \lgpgain{32.47}{1.2}
 & 31.64 & \lgpgain{32.17}{1.7}
 & 37.13 & \lgpgain{37.66}{1.4} \\
\midrule
\multirow{5}{*}{
  \makecell[l]{
    M-BEIR\\[-1pt]
    \footnotesize\textcolor{exactGray}{(16.16 / 22.87)}
  }
}
 & 10
 & 9.08 & \lgpgain{9.47}{4.3}
 & 13.07 & \lgpgain{13.69}{4.7}
 & 11.00 & \lgpgain{11.49}{4.5}
 & 13.07 & \lgpgain{13.69}{4.7} \\
 & 20
 & 10.29 & \lgpgain{11.49}{11.7}
 & 14.82 & \lgpgain{16.46}{11.1}
 & 13.90 & \lgpgain{15.73}{13.2}
 & 17.01 & \lgpgain{19.02}{11.8} \\
 & 50
 & 12.99 & \lgpgain{13.68}{5.3}
 & 18.28 & \lgpgain{19.19}{5.0}
 & 19.00 & \lgpgain{20.17}{6.2}
 & 22.81 & \lgpgain{24.01}{5.3} \\
 & 100
 & 14.00 & \lgpgain{14.66}{4.7}
 & 19.60 & \lgpgain{20.47}{4.4}
 & 20.93 & \lgpgain{22.00}{5.1}
 & 25.66 & \lgpgain{27.08}{5.5} \\
 & 200
 & 14.82 & \lgpgain{15.23}{2.8}
 & 20.62 & \lgpgain{21.25}{3.1}
 & 22.20 & \lgpgain{22.68}{2.2}
 & 27.30 & \lgpgain{27.77}{1.7} \\
\bottomrule
\end{tabular}
}
\vspace{-14pt}
\end{table*}

\noindent\textbf{Main Results.}
As shown in~\Cref{tab:bright-main}, \sys improves both greedy-search and reranked NDCG@10 across all 12 BRIGHT datasets at the search width of 20.
Under greedy search, \sys increases average NDCG@10 from 17.12 to 21.30 (\textbf{+24.4\%}) and Recall@10 from 20.01 to 25.05 (\textbf{+25.2\%}).
Biology shows the largest absolute NDCG@10 gain, increasing from 24.69 to 37.06 (\textbf{+50.1\%}).
These gains persist after LLM reranking, where average NDCG@10 increases from 18.92 to 23.83 (\textbf{+26.0\%}) and Recall@10 from 20.95 to 26.68 (\textbf{+27.4\%}).
%
As shown in~\Cref{tab:mbeir-main}, across 16 M-BEIR tasks, \sys improves average NDCG@10 by \textbf{11.7\%} under greedy search and \textbf{13.2\%} after VLM reranking, with reranked Recall@10 improving by \textbf{11.8\%}.
WebQA Task~2 shows the largest NDCG@10 gains, improving by \textbf{51.6\%} under greedy search and \textbf{44.1\%} after reranking.
These results show that \sys refines the index with connections that better reflect semantic relationships, improving graph-search retrieval and downstream reranking across both text and multimodal settings.

\begin{figure}[t]
\centering
\includegraphics[width=\linewidth]{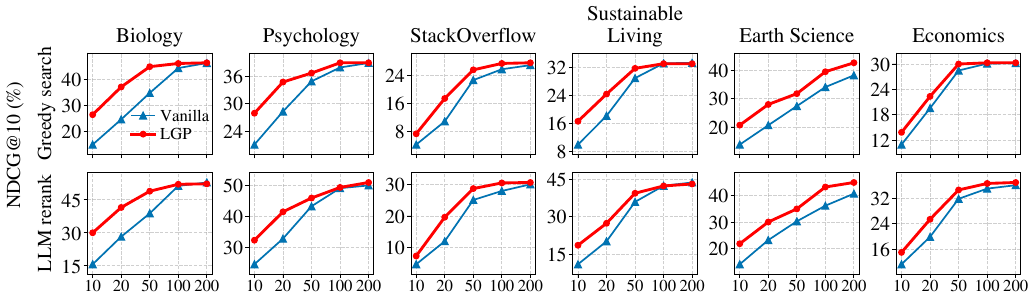}
\caption{
Search-width sweeps on 6 BRIGHT datasets using Diver embeddings.
\sys consistently outperforms Vanilla under both greedy search and LLM reranking across search widths.
}
\label{fig:bright-selected}
\vspace{-8pt}
\end{figure}


\begin{table*}[t]
\centering
\small
\setlength{\tabcolsep}{2.pt}
\renewcommand{\arraystretch}{1.04}
\caption{
Distance-computation-budget sweeps on BRIGHT using Diver embeddings.
}
\label{tab:computation-budget-main}
\scalebox{0.9}
{
\begin{tabular}{lccccccccc}
\toprule
\multirow{3}{*}[-1ex]{
\makecell[c]{
Benchmark\\[-1pt]
\footnotesize\textcolor{exactGray}{(Exact $k$NN ref.)}
}
}
& \multirow{3}{*}[-1ex]{\makecell[c]{Computation\\budget}}
& \multicolumn{4}{c}{Greedy search}
& \multicolumn{4}{c}{Rerank} \\
\cmidrule(lr){3-6}\cmidrule(lr){7-10}
& & \multicolumn{2}{c}{NDCG@10}
& \multicolumn{2}{c}{Recall@10}
& \multicolumn{2}{c}{NDCG@10}
& \multicolumn{2}{c}{Recall@10} \\
\cmidrule(lr){3-4}\cmidrule(lr){5-6}
\cmidrule(lr){7-8}\cmidrule(lr){9-10}
& & Vanilla & \sys
& Vanilla & \sys
& Vanilla & \sys
& Vanilla & \sys \\
\midrule
\multirow{7}{*}{
\makecell[l]{
BRIGHT\\[-1pt]
\footnotesize\textcolor{exactGray}{(28.50\,/\,34.18)}
}
}
& 100
& 3.96 & \lgpgain{5.02}{26.9}
& 2.96 & \lgpgain{4.15}{40.0}
& 4.28 & \lgpgain{5.40}{26.1}
& 3.07 & \lgpgain{4.12}{34.3} \\
& 200
& 10.50 & \lgpgain{13.98}{33.2}
& 11.62 & \lgpgain{15.60}{34.3}
& 11.67 & \lgpgain{15.41}{32.0}
& 12.32 & \lgpgain{16.40}{33.1} \\
& 300
& 16.69 & \lgpgain{19.80}{18.6}
& 19.18 & \lgpgain{23.19}{20.9}
& 18.70 & \lgpgain{22.46}{20.1}
& 20.35 & \lgpgain{24.61}{21.0} \\
& 500
& 22.92 & \lgpgain{24.57}{7.2}
& 27.14 & \lgpgain{28.83}{6.2}
& 26.04 & \lgpgain{27.79}{6.7}
& 28.70 & \lgpgain{30.82}{7.4} \\
& 800
& 25.89 & \lgpgain{26.61}{2.8}
& 30.66 & \lgpgain{31.37}{2.3}
& 29.00 & \lgpgain{29.92}{3.2}
& 32.59 & \lgpgain{33.54}{2.9} \\
& 1600
& 27.11 & \lgpgain{27.27}{0.6}
& 32.27 & \lgpgain{32.32}{0.2}
& 30.83 & \lgpgain{30.95}{0.4}
& 34.63 & \lgpgain{34.71}{0.2} \\
& 3200
& 27.65 & \lgpgain{27.74}{0.3}
& 32.90 & \lgpgain{32.96}{0.2}
& 31.26 & \lgpgain{31.28}{0.1}
& 35.15 & \lgpgain{35.18}{0.1} \\
\bottomrule
\end{tabular}
}
\vspace{-5pt}
\end{table*}

\begin{table*}[!t]
\centering
\small
\setlength{\tabcolsep}{2pt}
\renewcommand{\arraystretch}{1.02}
\caption{Cross-index evaluation of \sys on HNSW over BRIGHT datasets with Diver embeddings at search width 20. Each entry reports Vanilla\,/\,\sys.}
\label{tab:hnsw-main}
\resizebox{\textwidth}{!}{%
\begin{tabular}{lcccccccc}
\toprule
 & \multicolumn{4}{c}{Greedy search} & \multicolumn{4}{c}{Rerank} \\
\cmidrule(lr){2-5}\cmidrule(lr){6-9}
Dataset & NDCG@5 & NDCG@10 & Recall@5 & Recall@10 & NDCG@5 & NDCG@10 & Recall@5 & Recall@10 \\
\midrule
TheoremQA-Theorems & 38.64\,/\,\textbf{39.47} & 42.39\,/\,\textbf{43.34} & 45.05\,/\,\textbf{45.71} & 54.81\,/\,\textbf{56.12} & 38.26\,/\,\textbf{40.76} & 41.85\,/\,\textbf{43.52} & 44.86\,/\,\textbf{48.15} & 55.47\,/\,\textbf{56.12} \\
Biology & \textbf{42.71}\,/\,42.19 & \textbf{48.06}\,/\,47.39 & \textbf{42.13}\,/\,41.40 & \textbf{57.25}\,/\,56.09 & 50.85\,/\,\textbf{52.83} & 54.67\,/\,\textbf{55.74} & 48.63\,/\,\textbf{49.20} & \textbf{59.90}\,/\,58.90 \\
Pony & 14.84\,/\,\textbf{17.11} & 12.83\,/\,\textbf{14.48} & 3.83\,/\,\textbf{4.36} & 6.71\,/\,\textbf{8.04} & 17.82\,/\,\textbf{19.92} & 14.80\,/\,\textbf{16.28} & 4.72\,/\,\textbf{5.62} & 7.37\,/\,\textbf{8.78} \\
\bottomrule
\end{tabular}%
}
\vspace{-18pt}
\end{table*}


\begin{table}[t]
\centering
\small
\setlength{\tabcolsep}{4pt}
\renewcommand{\arraystretch}{1.08}
\caption{
Embedding robustness on BRIGHT using Qwen3-Embedding at search width 20.
Results are averaged equally over all 12 datasets.
}
\label{tab:bright-embedding}
{
\begin{tabular}{lcccccccc}
\toprule
& \multicolumn{2}{c}{NDCG@5}
& \multicolumn{2}{c}{NDCG@10}
& \multicolumn{2}{c}{Recall@5}
& \multicolumn{2}{c}{Recall@10} \\
\cmidrule(lr){2-3}
\cmidrule(lr){4-5}
\cmidrule(lr){6-7}
\cmidrule(lr){8-9}
Retrieval
& Vanilla & \sys
& Vanilla & \sys
& Vanilla & \sys
& Vanilla & \sys \\
\midrule
Greedy search
& 10.89 & \lgpgain{12.33}{13.2}
& 12.04 & \lgpgain{13.81}{14.7}
& 11.52 & \lgpgain{13.06}{13.4}
& 15.34 & \lgpgain{17.60}{14.7} \\
LLM rerank
& 15.91 & \lgpgain{18.19}{14.3}
& 16.34 & \lgpgain{18.70}{14.4}
& 15.31 & \lgpgain{17.37}{13.5}
& 18.19 & \lgpgain{20.66}{13.6} \\
\bottomrule
\end{tabular}
}
\vspace{-3pt}
\end{table}

\begin{table*}[!t]
\centering
\small
\caption{
Ablations of model size (left) and graph degree (right) on Pony and Psychology using Diver embeddings at search width 20.
Entries report NDCG@10.
}
\label{tab:ablation-model-degree}

\begin{minipage}[t]{0.54\textwidth}
\vspace{0pt}
\centering
\setlength{\tabcolsep}{2.2pt}
\renewcommand{\arraystretch}{1.06}
\begin{tabular}{lccccc}
\toprule
& & \multicolumn{2}{c}{Greedy search}
& \multicolumn{2}{c}{LLM reranking} \\
\cmidrule(lr){3-4}\cmidrule(lr){5-6}
Dataset & \makecell{Model} & Vanilla & \sys & Vanilla & \sys \\
\midrule
Pony
& Qwen3-8B  & \textbf{17.08} & 17.00 & 19.54 & \textbf{19.83} \\
& Qwen3-32B & 17.08 & \textbf{17.80} & 19.54 & \textbf{20.07} \\
\midrule
Psychology
& Qwen3-8B  & 28.31 & \textbf{29.82} & 32.55 & \textbf{34.75} \\
& Qwen3-32B & 28.31 & \textbf{34.76} & 32.69 & \textbf{41.46} \\
\bottomrule
\end{tabular}
\end{minipage}
\hfill
\begin{minipage}[t]{0.45\textwidth}
\vspace{0pt}
\centering
\setlength{\tabcolsep}{2pt}
\scalebox{0.85}{
\begin{tabular}{lccccc}
\toprule
& & \multicolumn{2}{c}{Greedy search}
& \multicolumn{2}{c}{LLM reranking} \\
\cmidrule(lr){3-4}\cmidrule(lr){5-6}
Dataset & $R$ & Vanilla & \sys & Vanilla & \sys \\
\midrule
Pony
& 8  & 5.08  & \textbf{9.80}  & 5.08  & \textbf{10.08} \\
& 16 & 17.08 & \textbf{17.80} & 19.54 & \textbf{20.07} \\
& 32 & 15.87 & \textbf{16.28} & 18.97 & \textbf{19.52} \\
\midrule
Psychology
& 8  & 6.74  & \textbf{18.52} & 7.71  & \textbf{22.68} \\
& 16 & 28.31 & \textbf{34.76} & 32.69 & \textbf{41.46} \\
& 32 & \textbf{37.66} & 36.22 & \textbf{45.17} & 43.78 \\
\bottomrule
\end{tabular}
}
\end{minipage}
\vspace{-5pt}
\end{table*}

\noindent\textbf{Search Width.}
To assess how the benefits of \sys vary with query-time search width, we evaluate performance across different widths.
\Cref{tab:search-width} shows that \sys achieves its largest gains on BRIGHT at narrower widths: at width 10, average NDCG@10 improves by \textbf{32.3\%} under greedy search and \textbf{32.8\%} after LLM reranking, while remaining consistently better than Vanilla as the search width increases. 
The per-dataset curves in~\Cref{fig:bright-selected} exhibit the same pattern.
On M-BEIR, \sys improves average NDCG@10 under both greedy search and reranking at every tested width.
The task-level results in~\Cref{fig:mbeir-selected} (Appendix) further show consistent gains across individual M-BEIR tasks.
Furthermore, compared with Vanilla, \sys approaches the exact $k$NN faster as the search width increases and remains consistently closer to exact retrieval on both benchmarks.
These results show that \sys provides its strongest gains at narrower search widths while remaining consistently beneficial as the search width increases.
Dataset-wise sweeps are provided in~\Cref{app:search-width-sweeps}.

\noindent\textbf{Explicit Distance-Computation Budgets.}
Search width is the standard control for query-time search effort, while explicit distance-computation budgets provide a complementary matched-compute view.
We therefore compare Vanilla and \sys under the same limit on the number of query--document distance computations.
As shown in~\Cref{tab:computation-budget-main}, \sys consistently outperforms Vanilla across all tested budgets, with the largest gains at smaller and intermediate budgets.
At a budget of 200, \sys improves greedy NDCG@10 and Recall@10 by \textbf{33.2\%} and \textbf{34.3\%}, respectively.
Moreover, on Economics from BRIGHT, \sys reaches the same greedy NDCG@10 as Vanilla using roughly 50\% fewer distance computations (800 vs.\ 1,600).
These results demonstrate that \sys improves search efficiency by achieving higher retrieval quality for the same computation and comparable quality with \emph{substantially fewer distance evaluations}.
Additional results are provided in~\Cref{app:comp}.

\noindent\textbf{Graph-Based ANN Algorithms.}
To evaluate whether \sys generalizes beyond DiskANN, we additionally instantiate it on HNSW across three representative BRIGHT retrieval settings: Biology (domain-specific QA), Pony (code retrieval), and TheoremQA-Theorems (theorem retrieval).
At search width 20, \Cref{tab:hnsw-main} shows consistent improvements after reranking across datasets.
Under greedy search, \sys improves TheoremQA-Theorems and Pony, while remaining comparable to baseline on Biology.
The search-width sweeps in~\Cref{fig:hnsw-finegrained} (Appendix) show that these gains generally persist across HNSW search widths, with larger improvements at narrower widths.
These results demonstrate that \sys generalizes across different graph-based ANN indices.

\noindent\textbf{Embeddings.}
To evaluate robustness to the embedding model, we replace Diver with Qwen3-Embedding.
As shown in~\Cref{tab:bright-embedding}, \sys improves every aggregate retrieval metric under both greedy search and LLM reranking.
Average NDCG@10 improves by \textbf{14.7\%} under greedy search and \textbf{14.4\%} after reranking.
These results show that the gains of \sys remain robust across different embedding models.
Additional per-dataset results are provided in~\Cref{app:embed}.

\noindent\textbf{Model Size and Graph Degree.}
We further study the effects of model size and graph degree.
As shown in~\Cref{tab:ablation-model-degree}, using Qwen3-32B for semantic selection consistently yields stronger results than Qwen3-8B.
Across graph degrees, \sys outperforms Vanilla in most settings, with consistent gains on Pony and substantial improvements on Psychology in several configurations.
Notably, \sys substantially improves both datasets at $R=8$.
These results show that stronger models can improve graph construction and that the benefits of \sys extend across different graph-degree settings.

\begin{figure}[!t]
\centering
\includegraphics[width=\linewidth]{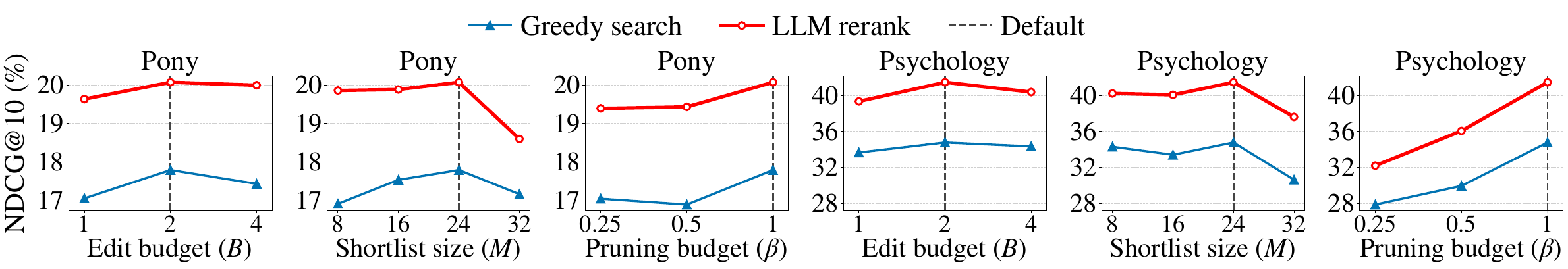}
\caption{
Hyperparameter ablation of \sys on Pony and Psychology at search width 20.
}
\label{fig:ablation-sensitivity}
\vspace{-15pt}
\end{figure}

\noindent\textbf{Per-Node Edit Budget, Shortlist Size, and Global Pruning Budget.}
We further evaluate \sys under different per-node edit budgets $B$, shortlist sizes $M$, and global pruning budgets $\beta$.
Here, $\beta$ denotes the maximum fraction of graph nodes that undergo LLM-guided pruning, with $\beta\in\{0.25,0.5,1\}$ corresponding to at most 25\%, 50\%, and 100\% of the graph, respectively.
As shown in~\Cref{fig:ablation-sensitivity}, the moderate local settings $B=2$ and $M=24$ achieve the strongest greedy-search and reranked NDCG@10 among the tested choices.
In contrast, increasing the global pruning budget improves performance, with $\beta=1$ achieving the strongest results on both Pony and Psychology.
These results suggest a clear trade-off: local updates should remain selective, since too few edits or candidates may limit semantic coverage, while overly aggressive editing or excessively large shortlists may hurt geometric structure or LLM selection quality.
At the graph level, however, applying LLM-guided pruning to more neighborhoods yields stronger retrieval performance.

\noindent\textbf{Additional Results.}
Additional search-width and distance-computation-budget sweeps, embedding-model, design, and hyperparameter ablations, and pruning-cost analysis are provided in~\Cref{app:additional}.

\section{Conclusion and Limitations}
We identify a fundamental \emph{geometry--semantic mismatch} in graph-based ANNS: indices are built using embedding-space geometry but evaluated by semantic relevance.
Our analysis shows that geometrically comparable edge choices can yield substantially different retrieval performance, motivating \sys, an offline framework that uses LLM reasoning to replace structurally low-value neighbors with semantically more useful alternatives.
Across textual and multimodal benchmarks, \sys consistently improves both greedy graph search and downstream reranking on DiskANN and HNSW while preserving graph sparsity and leaving the query-time search procedure unchanged.

Our current framework is limited to graph-based ANN indices.
Its effectiveness also depends on the quality of LLM/VLM semantic judgments.
Extending this index-level refinement paradigm to other ANN algorithms is a promising direction for future work.
More broadly, we view \sys as a first step towards identifying and addressing the geometry--semantic mismatch rather than the end of the story.

\newpage

\bibliography{src/references}
\bibliographystyle{iclr2027_conference}

\newpage

\appendix

\begin{table}[t]
\centering
\small
\setlength{\tabcolsep}{4pt}
\renewcommand{\arraystretch}{1.08}
\caption{Dataset statistics for the BRIGHT benchmark.}
\label{tab:bright-populations}
\begin{tabular}{lrr}
\toprule
Dataset & Queries & Documents \\
\midrule
Biology & 103 & 57,359 \\
Earth Science & 116 & 121,249 \\
Economics & 103 & 50,220 \\
Psychology & 101 & 52,835 \\
Robotics & 101 & 61,961 \\
StackOverflow & 117 & 107,081 \\
Sustainable Living & 108 & 60,792 \\
Pony & 112 & 7,894 \\
LeetCode & 142 & 413,932 \\
AoPS & 111 & 188,002 \\
TheoremQA-Theorems & 76 & 23,839 \\
TheoremQA-Questions & 194 & 188,002 \\
\bottomrule
\end{tabular}
\end{table}
\begin{table}[t]
\centering
\small
\setlength{\tabcolsep}{5pt}
\renewcommand{\arraystretch}{1.08}
\caption{Task subsets, retrieval modalities, and evaluation set sizes for the M-BEIR benchmark. I and T denote image and text, respectively.}
\label{tab:mbeir-populations}
\begin{tabular}{lclrr}
\toprule
Dataset & Task & Query $\to$ target & Queries & Candidates \\
\midrule
VisualNews & 0 & $\mathrm{T} \to \mathrm{I}$ & 19,995 & 542,246 \\
MSCOCO & 0 & $\mathrm{T} \to \mathrm{I}$ & 24,809 & 5,000 \\
Fashion200K & 0 & $\mathrm{T} \to \mathrm{I}$ & 1,719 & 201,824 \\
WebQA & 1 & $\mathrm{T} \to \mathrm{T}$ & 2,455 & 544,457 \\
EDIS & 2 & $\mathrm{T} \to \mathrm{I}+\mathrm{T}$ & 3,241 & 1,047,067 \\
WebQA & 2 & $\mathrm{T} \to \mathrm{I}+\mathrm{T}$ & 2,511 & 403,196 \\
VisualNews & 3 & $\mathrm{I} \to \mathrm{T}$ & 20,000 & 537,568 \\
MSCOCO & 3 & $\mathrm{I} \to \mathrm{T}$ & 5,000 & 24,809 \\
Fashion200K & 3 & $\mathrm{I} \to \mathrm{T}$ & 4,889 & 61,707 \\
NIGHTS & 4 & $\mathrm{I} \to \mathrm{I}$ & 2,120 & 40,038 \\
OVEN & 6 & $\mathrm{I}+\mathrm{T} \to \mathrm{T}$ & 50,004 & 676,667 \\
InfoSeek & 6 & $\mathrm{I}+\mathrm{T} \to \mathrm{T}$ & 11,323 & 611,651 \\
CIRR & 7 & $\mathrm{I}+\mathrm{T} \to \mathrm{I}$ & 4,170 & 21,551 \\
FashionIQ & 7 & $\mathrm{I}+\mathrm{T} \to \mathrm{I}$ & 6,003 & 74,381 \\
OVEN & 8 & $\mathrm{I}+\mathrm{T} \to \mathrm{I}+\mathrm{T}$ & 14,741 & 335,135 \\
InfoSeek & 8 & $\mathrm{I}+\mathrm{T} \to \mathrm{I}+\mathrm{T}$ & 17,593 & 481,782 \\
\bottomrule
\end{tabular}
\end{table}

\begin{table}[t]
\centering
\small
\setlength{\tabcolsep}{5pt}
\renewcommand{\arraystretch}{1.08}
\caption{Models used in the experiments.}
\label{tab:experiment-models}
\begin{tabular}{lll}
\toprule
Benchmark & Role & Model \\
\midrule
BRIGHT & Embedding & Diver-Retriever-4B-1020 \\
BRIGHT & Embedding & Qwen3-Embedding-4B \\
BRIGHT & Graph refinement / reranking & Qwen3-32B \\
M-BEIR & Embedding & CLIP (ViT-B/32) \\
M-BEIR & Graph refinement / reranking & Qwen3-VL-30B-A3B-Instruct \\
\bottomrule
\end{tabular}
\end{table}

\section{Experimental Setup}\label{app:exp_setup}
\noindent\textbf{Benchmarks.}
We use BRIGHT~\citep{su2025bright} for reasoning-intensive text retrieval and M-BEIR~\citep{10.1007/978-3-031-73021-4_23} for multimodal retrieval.
BRIGHT contains 12 datasets spanning diverse domains, while M-BEIR contains 16 tasks covering text, image, and image--text retrieval.
For BRIGHT, we evaluate all queries in each dataset.
For M-BEIR, following the setup of~\citet{xu2026beyond}, we uniformly sample 100 queries from each task and use the same sampled queries across all experiments.
Following the task-specific M-BEIR preprocessing~\citep{10.1007/978-3-031-73021-4_23,xu2026beyond}, we use the released text corpora for OVEN and InfoSeek Task~6, restrict retrieval to their corresponding candidate pools, and use the associated relevance labels for evaluation.
Detailed benchmark statistics are provided in~\Cref{tab:bright-populations,tab:mbeir-populations}.

\noindent\textbf{Models.}
For BRIGHT, we use Diver-Retriever-4B-1020~\citep{sun2026divermultistageapproachreasoningintensive} and Qwen3-Embedding-4B as embedding models.
For M-BEIR, following~\citet{xu2026beyond}, we use CLIP (ViT-B/32)~\citep{pmlr-v139-radford21a} to encode all 16 task subsets.
We follow the benchmark-specific query formatting by prepending each text-bearing query with its corresponding task instruction.
For single-modality inputs, we $\ell_2$-normalize the corresponding CLIP feature.
For image--text inputs, we first $\ell_2$-normalize the image and text features separately, sum the two normalized vectors, and then normalize the resulting vector again.
Retrieval is performed using cosine similarity.
For LLM-guided semantic selection in \sys and downstream reranking, we use Qwen3-32B~\citep{qwen3technicalreport} for BRIGHT and Qwen3-VL-30B-A3B-Instruct~\citep{bai2025qwen3vltechnicalreport} for M-BEIR.
Both models are used in their non-reasoning configuration.
For the model-size ablation, we vary only the model used for semantic selection, using Qwen3-8B or Qwen3-32B, while keeping Qwen3-32B fixed as the downstream reranker.
\Cref{tab:experiment-models} summarizes the specific model configurations.

\noindent\textbf{Index Configurations.}
Our primary ANN index is DiskANN~\citep{NEURIPS2019_09853c7f}, configured with a maximum out-degree of 16, a construction search-list size of 125, and a pruning parameter of 1.2.
We additionally evaluate HNSW~\citep{10.1109/TPAMI.2018.2889473}, using a connectivity parameter of 16 and a construction search width of 200.
For both indices, \sys performs local graph pruning on top of the corresponding Vanilla index.
For the textual retrieval setting, \sys uses edit budget $B=2$, shortlist size $M=24$, neighborhood-context size $|K_i|=4$, and threshold $g_{\min}=12$.
For the multimodal setting, we use $B=2$ and $M=12$, with the same context size and threshold.
For the graph-degree ablation, we set $g_{\min}=6$ at $R=8$ to match the default $g_{\min}/R$ ratio at $R=16$, while using $g_{\min}=12$ for all other degree settings.

\noindent\textbf{Retrieval Configurations and Metrics.}
At query time, we evaluate two settings: (i) greedy search, and (ii) downstream LLM/VLM reranking of the candidate set returned by greedy search.
We use \emph{search width}, the standard query-time search-effort parameter in graph-based ANN indices, corresponding to $L_{\mathrm{search}}$ in DiskANN and $ef_{\mathrm{search}}$ in HNSW.
Search width controls the breadth of candidates explored during graph traversal, with larger values generally requiring more node expansions and distance computations.
With graph degree bounded by $R$, the distance-computation cost grows roughly with $\text{search width}\times R$, making search width a practical control for query-time search effort.
We therefore use search width as the primary query-time search-effort parameter throughout our main evaluation.
For DiskANN, we vary $L_{\mathrm{search}}\in\{10,20,50,100,200\}$; for HNSW, we vary $ef_{\mathrm{search}}\in\{10,12,14,16,18,20,50\}$.
By default, we set the search width to 20 for both indices to target the budget-constrained regime, where graph connectivity most strongly dictates which relevant documents are reached.
For reranking, we use the candidates returned by graph search, capped at 200 for text retrieval and 64 for multimodal retrieval.
At the default search width of 20, both rerankers therefore receive 20 candidates.
Both rerankers use listwise sliding windows of 10 candidates with a stride of 5.
For greedy search, the top $k$ documents are selected from the returned candidates according to embedding distance.
For reranking, the same candidate set is reordered by the model before evaluating the resulting top $k$.

Our primary retrieval metrics are NDCG@$k$ and document-level Recall@$k$ for $k\in\{5,10\}$, reported on a 0--100 scale.
NDCG@$k$ measures ranking quality within the top $k$, normalized by the ideal ranking at the same cutoff.
Recall@$k$ measures the fraction of all ground-truth documents that appear in the top $k$, averaged over queries.
Because reranking can change which documents appear in the top $k$, it can also change Recall@$k$ even though the candidate set itself is unchanged.
When a query has more than $k$ relevant documents, its maximum Recall@$k$ is below 100, so Recall@$k$ and NDCG@$k$ are not directly comparable in magnitude.

\begin{table*}[t]
\centering
\small
\setlength{\tabcolsep}{2pt}
\renewcommand{\arraystretch}{1.02}
\caption{BRIGHT main results on DiskANN at search width 20 using Diver embeddings. Each entry reports Vanilla\,/\,\sys.}
\label{tab:bright-main-app}
\resizebox{\textwidth}{!}
{%
\begin{tabular}{lcccccccc}
\toprule
 & \multicolumn{4}{c}{Greedy search} & \multicolumn{4}{c}{Rerank} \\
\cmidrule(lr){2-5}\cmidrule(lr){6-9}
Dataset & NDCG@5 & NDCG@10 & Recall@5 & Recall@10 & NDCG@5 & NDCG@10 & Recall@5 & Recall@10 \\
\midrule
Biology & 22.22\,/\,\textbf{32.66} & 24.69\,/\,\textbf{37.06} & 21.42\,/\,\textbf{32.10} & 28.66\,/\,\textbf{44.63} & 26.95\,/\,\textbf{39.15} & 28.30\,/\,\textbf{41.57} & 26.22\,/\,\textbf{38.47} & 30.71\,/\,\textbf{46.19} \\
Earth Science & 19.80\,/\,\textbf{26.32} & 20.60\,/\,\textbf{27.92} & 19.13\,/\,\textbf{25.80} & 22.70\,/\,\textbf{30.59} & 21.91\,/\,\textbf{28.03} & 23.29\,/\,\textbf{30.06} & 20.62\,/\,\textbf{27.30} & 25.51\,/\,\textbf{33.03} \\
Economics & 18.68\,/\,\textbf{21.00} & 19.56\,/\,\textbf{22.35} & 15.62\,/\,\textbf{18.81} & 20.77\,/\,\textbf{25.61} & 18.84\,/\,\textbf{24.01} & 19.98\,/\,\textbf{25.42} & 16.09\,/\,\textbf{22.46} & 21.92\,/\,\textbf{28.84} \\
Psychology & 26.62\,/\,\textbf{32.37} & 28.31\,/\,\textbf{34.76} & 26.62\,/\,\textbf{31.62} & 34.51\,/\,\textbf{41.61} & 32.88\,/\,\textbf{40.20} & 32.78\,/\,\textbf{41.46} & 29.47\,/\,\textbf{35.11} & 34.40\,/\,\textbf{44.10} \\
Robotics & 19.05\,/\,\textbf{21.19} & 19.48\,/\,\textbf{21.68} & 20.59\,/\,\textbf{22.35} & 24.83\,/\,\textbf{27.12} & 25.68\,/\,\textbf{28.47} & 25.67\,/\,\textbf{28.63} & 24.27\,/\,\textbf{25.98} & 28.61\,/\,\textbf{31.16} \\
StackOverflow & 8.75\,/\,\textbf{14.83} & 10.82\,/\,\textbf{17.36} & 8.71\,/\,\textbf{14.76} & 13.87\,/\,\textbf{21.37} & 10.22\,/\,\textbf{16.86} & 11.96\,/\,\textbf{19.61} & 9.72\,/\,\textbf{15.94} & 14.66\,/\,\textbf{23.73} \\
Sustainable Living & 16.35\,/\,\textbf{22.30} & 18.09\,/\,\textbf{24.33} & 15.09\,/\,\textbf{22.12} & 22.12\,/\,\textbf{29.76} & 19.32\,/\,\textbf{26.05} & 20.20\,/\,\textbf{27.43} & 18.21\,/\,\textbf{25.06} & 22.04\,/\,\textbf{30.82} \\
Pony & 20.34\,/\,\textbf{20.99} & 17.08\,/\,\textbf{17.80} & 5.29\,/\,\textbf{5.36} & 9.06\,/\,\textbf{9.40} & 24.81\,/\,\textbf{26.04} & 19.54\,/\,\textbf{20.07} & 6.56\,/\,\textbf{6.73} & \textbf{9.73}\,/\,9.00 \\
LeetCode & 2.66\,/\,\textbf{4.89} & 3.20\,/\,\textbf{6.20} & 3.35\,/\,\textbf{6.27} & 4.54\,/\,\textbf{9.51} & 2.34\,/\,\textbf{5.27} & 2.97\,/\,\textbf{6.37} & 2.66\,/\,\textbf{7.05} & 4.27\,/\,\textbf{10.07} \\
AoPS & 2.67\,/\,2.67 & 3.13\,/\,\textbf{3.36} & 3.03\,/\,3.03 & 4.58\,/\,\textbf{5.03} & 2.39\,/\,\textbf{2.46} & 3.12\,/\,\textbf{3.23} & 2.52\,/\,2.52 & 4.69\,/\,\textbf{4.79} \\
TheoremQA-Theorems & 31.23\,/\,\textbf{32.63} & 34.70\,/\,\textbf{36.35} & 37.16\,/\,\textbf{37.48} & 46.26\,/\,\textbf{47.57} & 30.36\,/\,\textbf{33.18} & 33.57\,/\,\textbf{36.44} & 36.31\,/\,\textbf{39.27} & 45.60\,/\,\textbf{48.89} \\
TheoremQA-Questions & 5.39\,/\,\textbf{6.20} & 5.73\,/\,\textbf{6.45} & 7.56\,/\,\textbf{7.86} & 8.24\,/\,\textbf{8.36} & \textbf{4.92}\,/\,4.90 & 5.63\,/\,\textbf{5.68} & 7.52\,/\,\textbf{7.56} & 9.24\,/\,\textbf{9.49} \\
\midrule
Average & 16.15\,/\,\textbf{19.84} & 17.12\,/\,\textbf{21.30} & 15.30\,/\,\textbf{18.96} & 20.01\,/\,\textbf{25.05} & 18.38\,/\,\textbf{22.88} & 18.92\,/\,\textbf{23.83} & 16.68\,/\,\textbf{21.12} & 20.95\,/\,\textbf{26.68} \\
\bottomrule
\end{tabular}%
}
\end{table*}

\begin{table*}[t]
\centering
\small
\setlength{\tabcolsep}{2pt}
\renewcommand{\arraystretch}{1.02}
\caption{M-BEIR main results on DiskANN at search width 20 using CLIP (ViT-B/32). Parentheses indicate the M-BEIR task ID. Each entry reports Vanilla\,/\,\sys.}
\label{tab:mbeir-main-app}
\resizebox{\textwidth}{!}{%
\begin{tabular}{lcccccccc}
\toprule
 & \multicolumn{4}{c}{Greedy search} & \multicolumn{4}{c}{Rerank} \\
\cmidrule(lr){2-5}\cmidrule(lr){6-9}
Dataset (Task) & NDCG@5 & NDCG@10 & Recall@5 & Recall@10 & NDCG@5 & NDCG@10 & Recall@5 & Recall@10 \\
\midrule
VisualNews (0) & \textbf{15.80}\,/\,14.54 & \textbf{16.41}\,/\,15.45 & \textbf{19.00}\,/\,17.00 & \textbf{21.00}\,/\,20.00 & 14.26\,/\,\textbf{15.02} & 15.23\,/\,\textbf{15.97} & 18.00\,/\,\textbf{18.00} & 21.00\,/\,\textbf{21.00} \\
MSCOCO (0) & 28.26\,/\,\textbf{28.33} & 32.45\,/\,\textbf{32.88} & 38.00\,/\,\textbf{38.00} & 51.00\,/\,\textbf{52.00} & 42.72\,/\,\textbf{43.72} & 45.09\,/\,\textbf{46.09} & 49.00\,/\,\textbf{50.00} & 56.00\,/\,\textbf{57.00} \\
Fashion200K (0) & 1.51\,/\,\textbf{1.51} & 1.80\,/\,\textbf{1.82} & 1.79\,/\,\textbf{1.79} & 2.63\,/\,\textbf{2.63} & \textbf{4.41}\,/\,3.41 & \textbf{4.53}\,/\,3.56 & \textbf{3.97}\,/\,2.97 & \textbf{4.97}\,/\,3.97 \\
WebQA (1) & 5.38\,/\,\textbf{6.00} & 5.99\,/\,\textbf{6.61} & 6.00\,/\,\textbf{6.50} & 7.50\,/\,\textbf{8.00} & 8.29\,/\,\textbf{8.76} & 8.29\,/\,\textbf{8.93} & 8.00\,/\,\textbf{8.50} & 8.00\,/\,\textbf{9.00} \\
EDIS (2) & 5.87\,/\,\textbf{9.89} & 6.76\,/\,\textbf{11.29} & 6.49\,/\,\textbf{10.57} & 9.07\,/\,\textbf{14.74} & 10.33\,/\,\textbf{14.22} & 11.02\,/\,\textbf{15.67} & 10.63\,/\,\textbf{14.05} & 12.80\,/\,\textbf{18.71} \\
WebQA (2) & 12.99\,/\,\textbf{21.73} & 15.15\,/\,\textbf{22.97} & 14.50\,/\,\textbf{26.00} & 20.50\,/\,\textbf{29.50} & 19.50\,/\,\textbf{28.56} & 20.25\,/\,\textbf{29.18} & 21.50\,/\,\textbf{32.00} & 23.50\,/\,\textbf{33.50} \\
VisualNews (3) & 2.65\,/\,\textbf{5.77} & 3.87\,/\,\textbf{7.04} & 5.00\,/\,\textbf{10.00} & 9.00\,/\,\textbf{14.00} & 3.97\,/\,\textbf{7.77} & 5.25\,/\,\textbf{10.29} & 7.00\,/\,\textbf{11.00} & 11.00\,/\,\textbf{19.00} \\
MSCOCO (3) & 37.05\,/\,\textbf{37.51} & 42.42\,/\,\textbf{43.00} & 32.80\,/\,\textbf{32.80} & 42.60\,/\,\textbf{42.80} & 49.83\,/\,\textbf{51.11} & 52.63\,/\,\textbf{53.92} & 44.20\,/\,\textbf{44.80} & 49.20\,/\,\textbf{49.80} \\
Fashion200K (3) & \textbf{1.93}\,/\,1.86 & \textbf{2.93}\,/\,2.86 & 3.00\,/\,\textbf{3.00} & 6.00\,/\,\textbf{6.00} & 3.26\,/\,\textbf{3.65} & 3.97\,/\,\textbf{4.00} & 4.00\,/\,\textbf{5.00} & 6.00\,/\,\textbf{6.00} \\
NIGHTS (4) & 23.93\,/\,\textbf{23.93} & 28.73\,/\,\textbf{28.73} & 36.00\,/\,\textbf{36.00} & 51.00\,/\,\textbf{51.00} & 38.01\,/\,\textbf{39.22} & 41.99\,/\,\textbf{42.51} & 49.00\,/\,\textbf{51.00} & 61.00\,/\,\textbf{61.00} \\
OVEN (6) & 0.00\,/\,0.00 & 0.00\,/\,0.00 & 0.00\,/\,0.00 & 0.00\,/\,0.00 & 0.00\,/\,0.00 & 0.00\,/\,0.00 & 0.00\,/\,0.00 & 0.00\,/\,0.00 \\
InfoSeek (6) & 0.00\,/\,0.00 & 0.00\,/\,0.00 & 0.00\,/\,0.00 & 0.00\,/\,0.00 & 0.00\,/\,0.00 & 0.00\,/\,0.00 & 0.00\,/\,0.00 & 0.00\,/\,0.00 \\
CIRR (7) & 6.16\,/\,\textbf{6.86} & 6.47\,/\,\textbf{7.83} & 12.00\,/\,\textbf{13.00} & 13.00\,/\,\textbf{16.00} & 10.99\,/\,\textbf{12.34} & 10.99\,/\,\textbf{12.34} & 13.00\,/\,\textbf{15.00} & 13.00\,/\,\textbf{15.00} \\
FashionIQ (7) & 0.00\,/\,\textbf{0.43} & 0.67\,/\,\textbf{1.12} & 0.00\,/\,\textbf{1.00} & 2.00\,/\,\textbf{3.00} & 0.63\,/\,\textbf{2.63} & 1.61\,/\,\textbf{3.66} & 1.00\,/\,\textbf{3.00} & 4.00\,/\,\textbf{6.00} \\
OVEN (8) & 0.25\,/\,\textbf{0.59} & 0.50\,/\,\textbf{1.35} & 0.25\,/\,\textbf{0.27} & 0.75\,/\,\textbf{2.08} & 1.20\,/\,\textbf{3.39} & 1.16\,/\,\textbf{3.09} & 0.75\,/\,\textbf{2.08} & 0.75\,/\,\textbf{2.08} \\
InfoSeek (8) & 0.26\,/\,\textbf{0.41} & 0.48\,/\,\textbf{0.93} & 0.50\,/\,\textbf{0.58} & 1.00\,/\,\textbf{1.58} & 0.26\,/\,\textbf{2.82} & 0.45\,/\,\textbf{2.53} & 0.50\,/\,\textbf{1.62} & 1.00\,/\,\textbf{2.31} \\
\midrule
Average & 8.88\,/\,\textbf{9.96} & 10.29\,/\,\textbf{11.49} & 10.96\,/\,\textbf{12.28} & 14.82\,/\,\textbf{16.46} & 12.98\,/\,\textbf{14.79} & 13.90\,/\,\textbf{15.73} & 14.41\,/\,\textbf{16.19} & 17.01\,/\,\textbf{19.02} \\
\bottomrule
\end{tabular}%
}
\end{table*}

\begin{figure}[t]
\centering
\includegraphics[width=\linewidth]{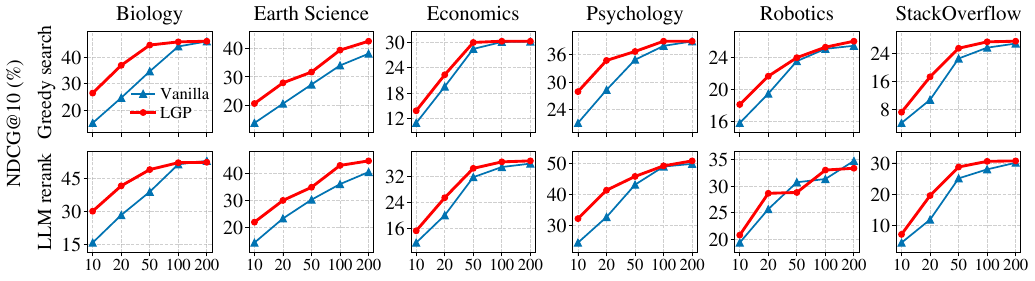}
\caption{Search-width sweeps on six BRIGHT datasets using Diver embeddings (part 1 of 2).}
\label{fig:bright-diver-1}
\end{figure}

\begin{figure}[t]
\centering
\includegraphics[width=0.95\linewidth]{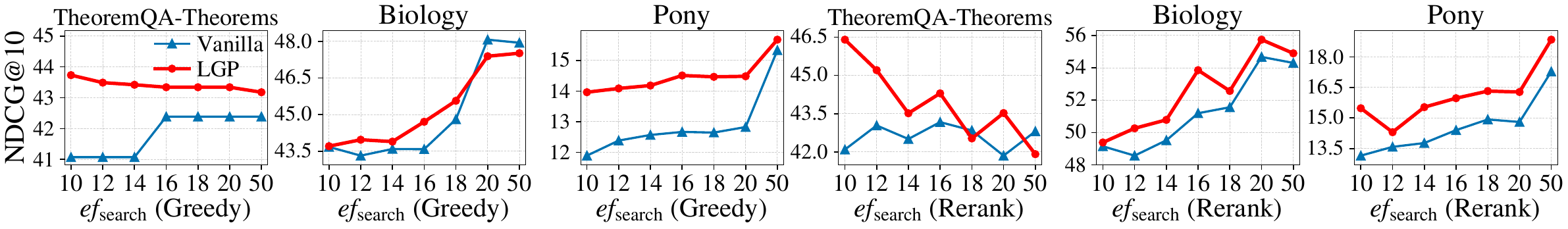}
\caption{HNSW search-width sweeps on 3 BRIGHT datasets. The left three panels show greedy-search NDCG@10, while the right three show NDCG@10 after LLM reranking.}
\label{fig:hnsw-finegrained}
\end{figure}

\begin{figure}[t]
\centering
\includegraphics[width=\linewidth]{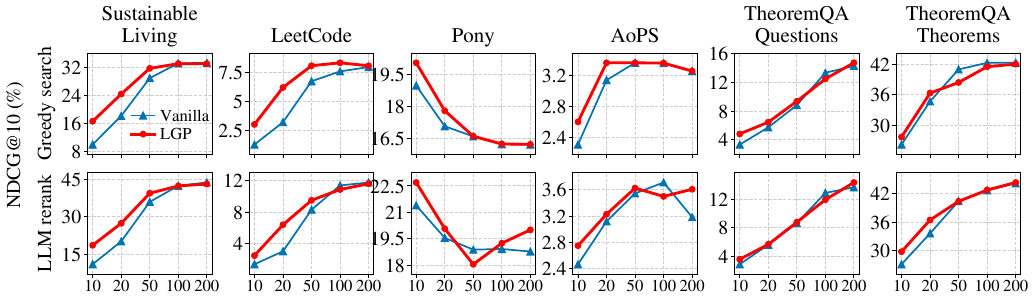}
\caption{Search-width sweeps on the remaining six BRIGHT datasets using Diver embeddings (part 2 of 2).}
\label{fig:bright-diver-2}
\end{figure}

\begin{figure}[t]
\centering
\includegraphics[width=0.95\linewidth]{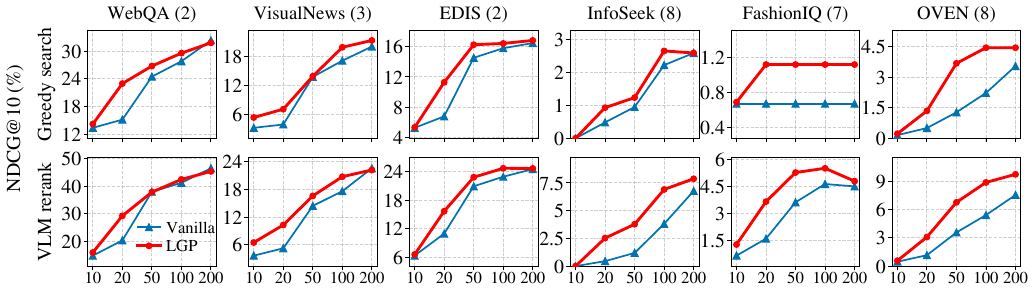}
\caption{
Search-width sweeps on 6 M-BEIR tasks using CLIP embeddings.
\sys consistently matches or outperforms Vanilla under both greedy search and VLM reranking across search widths.
}
\label{fig:mbeir-selected}
\end{figure}

\section{Additional Results}\label{app:additional}

\subsection{Search-Width Results}\label{app:search-width-sweeps}
\Cref{fig:bright-diver-1,fig:bright-diver-2} show the per-dataset search-width sweeps using Diver embeddings on BRIGHT.
The gains are broadly consistent across datasets, showing that the aggregate improvements are not driven by a small subset of tasks.
This further supports that \sys refines graph connectivity in ways that better align with downstream semantic relevance.

\begin{figure}[!t]
\centering
\includegraphics[width=\linewidth]{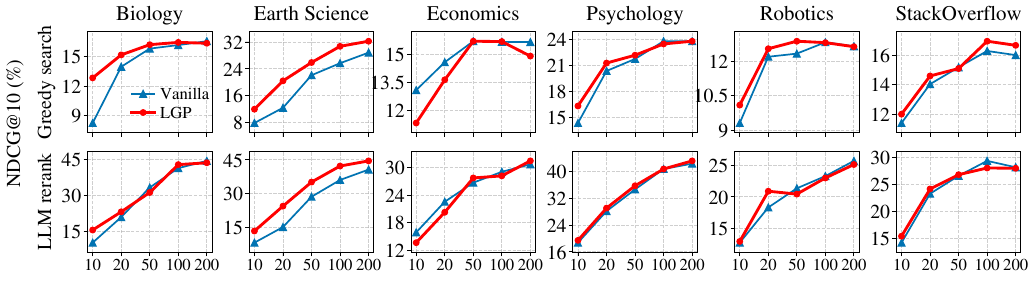}
\caption{Search-width sweeps on six BRIGHT datasets using Qwen3-Embedding (part 1 of 2).}
\label{fig:bright-qwen-1}
\end{figure}

\begin{figure}[t]
\centering
\includegraphics[width=\linewidth]{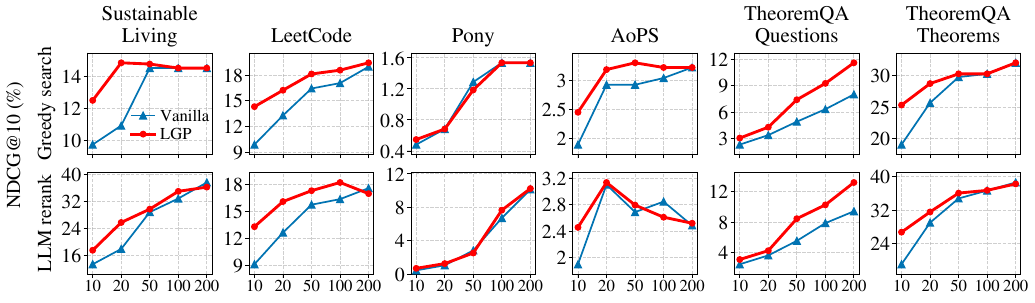}
\caption{Search-width sweeps on the remaining six BRIGHT datasets using Qwen3-Embedding (part 2 of 2).}
\label{fig:bright-qwen-2}
\end{figure}

\begin{table*}[!t]
\centering
\small
\renewcommand{\arraystretch}{1.06}
\caption{
Ablation of candidate shortlisting on Pony, Psychology, and Biology from BRIGHT at search width 20 using Diver embeddings.
Entries report NDCG@10.
}
\label{tab:ablation-shortlisting}
\begin{tabular}{lcccccc}
\toprule
& \multicolumn{2}{c}{Pony}
& \multicolumn{2}{c}{Psychology}
& \multicolumn{2}{c}{Biology} \\
\cmidrule(lr){2-3}
\cmidrule(lr){4-5}
\cmidrule(lr){6-7}
Variant & Greedy & Rerank & Greedy & Rerank & Greedy & Rerank \\
\midrule
Proximity only  & 16.89 & 18.94 & 33.71 & 39.52 & 33.07 & 37.69 \\
Coverage only   & \textbf{17.84} & \textbf{20.22} & 33.18 & 39.30 & 37.06 & 41.17 \\
Separation only & 17.19 & 20.02 & 33.45 & 39.68 & \textbf{39.22} & \textbf{45.71} \\
Balanced       & 17.80 & 20.07 & \textbf{34.76} & \textbf{41.46} & 37.06 & 41.57 \\
\bottomrule
\end{tabular}
\end{table*}

\begin{table*}[!t]
\centering
\small
\renewcommand{\arraystretch}{1.06}
\caption{
Ablation of low-value neighbor identification on Pony, Psychology, and Biology from BRIGHT at search width 20 using Diver embeddings.
Entries report NDCG@10.
}
\label{tab:ablation-removal}
\begin{tabular}{lcccccc}
\toprule
& \multicolumn{2}{c}{Pony}
& \multicolumn{2}{c}{Psychology}
& \multicolumn{2}{c}{Biology} \\
\cmidrule(lr){2-3}
\cmidrule(lr){4-5}
\cmidrule(lr){6-7}
Variant & Greedy & Rerank & Greedy & Rerank & Greedy & Rerank \\
\midrule
Redundancy only & 17.36 & \textbf{20.22} & 32.38 & 38.04 & 33.93 & 39.74 \\
Coverage only   & 17.73 & 20.03 & 34.30 & 41.15 & 34.41 & 38.86 \\
Lexicographic   & \textbf{17.80} & 20.07 & \textbf{34.76} & \textbf{41.46} & \textbf{37.06} & \textbf{41.57} \\
\bottomrule
\end{tabular}
\end{table*}




\subsection{Embedding-Model Ablations}\label{app:embed}
Per-dataset BRIGHT results using Qwen3-Embedding are shown in~\Cref{fig:bright-qwen-1,fig:bright-qwen-2}.
Across datasets and search widths, \sys consistently matches or outperforms Vanilla, further demonstrating that its gains are robust to the choice of embedding model.


\begin{table}[!t]
\centering
\small
\renewcommand{\arraystretch}{1.08}
\caption{
Graph-structure diagnostics for paired Vanilla and \sys indices on BRIGHT.
Each dataset contributes equally to the reported averages, and all indices use a maximum degree of 16.
Changed neighborhoods denote the percentage of nodes whose outgoing-neighbor sets differ between the paired indices.
}
\label{tab:index-diagnostics}
\begin{tabular}{lccc}
\toprule
\multicolumn{1}{c}{\multirow{2}{*}{Embedding}}
& \multicolumn{2}{c}{Mean out-degree}
& \multirow{2}{*}{Changed neighborhoods (\%)} \\
\cmidrule(lr){2-3}
& Vanilla & \sys & \\
\midrule
Diver           & 15.60 & 15.66 & 88.5 \\
Qwen3-Embedding & 15.88 & 15.80 & 88.9 \\
\bottomrule
\end{tabular}
\end{table}

\subsection{Design Analysis of \sys}

\noindent\textbf{Candidate Shortlisting.}
We ablate the shortlisting strategy while keeping the other designs and parameters of \sys fixed.
We consider three alternative shortlisting methods:
(i) \emph{Proximity only} selects candidates closest to the anchor in embedding space after excluding those geometrically covered by its current neighbors;
(ii) \emph{Coverage only} ranks candidates by decreasing additional two-hop coverage; and
(iii) \emph{Separation only} ranks candidates by decreasing minimum embedding distance to the current neighbors.
The default \emph{Balanced} strategy in \sys combines candidates from all three rankings under the same shortlist budget.
As shown in~\Cref{tab:ablation-shortlisting}, the most effective shortlisting strategy varies across datasets, with coverage favored on Pony, balanced on Psychology, and separation on Biology.
This variation exactly motivates our balanced strategy, which combines all three signals and maintains strong performance across datasets and retrieval settings.
Thus, balanced shortlisting provides a robust default that generalizes across datasets without dataset-specific tuning.

\noindent\textbf{Low-Value Neighbor Identification.}
We ablate the criteria used to identify low-value existing neighbors for replacement while keeping all other components of \sys fixed.
Besides our default \emph{Lexicographic} rule, we consider two additional variants:
(i) \emph{Redundancy only}, which favors geometrically redundant neighbors, and
(ii) \emph{Coverage only}, which identifies neighbors with the smallest unique contribution to two-hop coverage.
The default \emph{Lexicographic} rule first prioritizes low unique coverage, then greater redundancy.
As shown in~\Cref{tab:ablation-removal}, the lexicographic rule achieves the strongest greedy-search performance on all three datasets and the strongest reranked performance on Psychology and Biology, while remaining close to the best result on Pony.
These results show that coverage and redundancy provide complementary signals for identifying low-value neighbors, making the lexicographic rule a robust default across datasets.

\noindent\textbf{Graph Structure After Pruning.}
As shown in~\Cref{tab:index-diagnostics}, \sys changes approximately 88--89\% of neighborhoods across BRIGHT, while the mean out-degree remains nearly unchanged for both embedding models and stays below the maximum degree of 16.
Thus, the retrieval gains of \sys come from reconfiguring graph connectivity rather than increasing graph density.

\begin{figure}[t]
\centering
\includegraphics[width=0.8\linewidth]{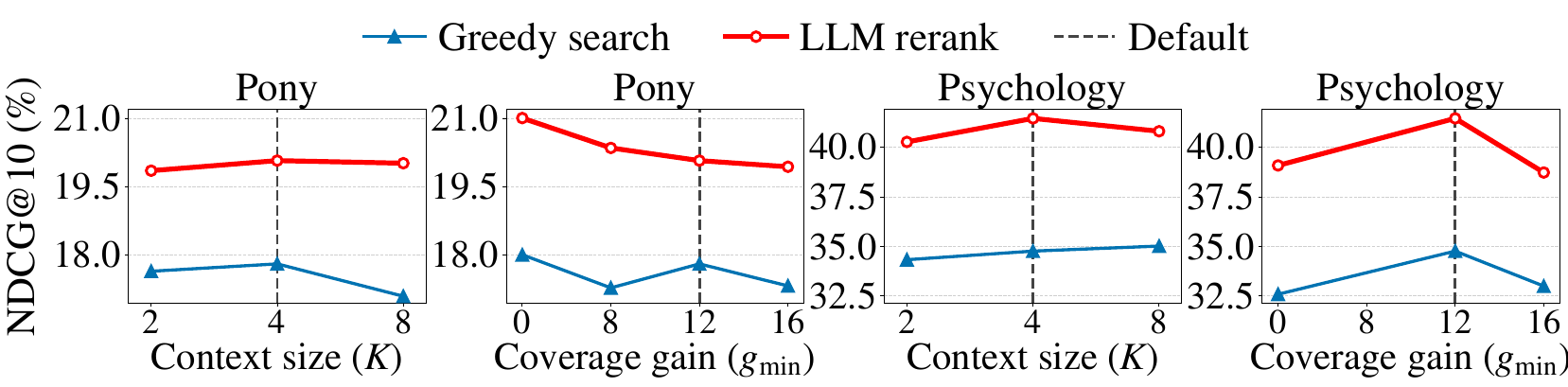}
\caption{
Additional hyperparameter ablations at search width 20.
The left two panels show Pony and the right two show Psychology, with dashed lines indicating the default settings.
}
\label{fig:ablation-sensitivity-appendix}
\end{figure}

\begin{table}[!t]
\centering
\small
\renewcommand{\arraystretch}{1.08}
\caption{
Offline graph-construction cost using Diver embeddings and Qwen3-32B on a single NVIDIA H200.
Build times are reported in minutes, token counts in millions, and per-node time as total \sys build time divided by the number of graph nodes.
LLM calls and token counts refer to \sys.
}
\label{tab:pruning-cost}
\begin{tabular}{@{}lrrrrrrr@{}}
\toprule
\multirow{2}{*}{Dataset}
& \multirow{2}{*}{Nodes}
& \multicolumn{2}{c}{Build time (min)}
& \multirow{2}{*}{\makecell[c]{\sys time\\(s/node)}}
& \multirow{2}{*}{LLM calls}
& \multicolumn{2}{c}{Tokens (M)} \\
\cmidrule(lr){3-4}\cmidrule(lr){7-8}
& & Vanilla & \sys & & & Input & Output \\
\midrule
Pony       & 7,894  & 1.49  & 109.46 & 0.83 & 7,894  & 23.95  & 0.22 \\
Psychology & 52,835 & 11.84 & 691.40 & 0.79 & 52,835 & 124.56 & 1.50 \\
\bottomrule
\end{tabular}
\end{table}

\subsection{Additional Ablation on Hyperparameters in \sys}
\noindent\textbf{Neighborhood Context Size $K$.}
We vary the neighborhood context size $K$, which controls the number of current neighbors provided to the LLM/VLM when selecting new neighbors that add complementary semantic coverage.
As shown in~\Cref{fig:ablation-sensitivity-appendix}, the default $K=4$ gives the strongest reranked NDCG@10 on both Pony and Psychology.
Increasing the context to $K=8$ provides no additional reranking benefit, suggesting that a small set of representative neighbors already captures most of the useful neighborhood context, while additional neighbors may introduce redundant information that makes semantic distinctions harder for the LLM/VLM.

\noindent\textbf{Coverage Threshold $g_{\min}$.}
We vary the coverage threshold $g_{\min}$, which controls the minimum additional two-hop coverage required for an LLM-selected candidate to be accepted as a new neighbor.
As shown in~\Cref{fig:ablation-sensitivity-appendix}, its effect varies across datasets: Pony performs best with the relaxed setting $g_{\min}=0$, whereas Psychology performs best at the default $g_{\min}=12$ under both greedy search and reranking.
These results indicate that the preferred coverage constraint is dataset-dependent rather than uniformly favoring stronger coverage requirements.
As shown in~\Cref{fig:ablation-sensitivity-appendix}, the default $g_{\min}=12$ remains competitive on Pony and achieves the strongest performance on Psychology under both greedy search and reranking.

\subsection{Pruning Cost Analysis}
We profile the offline construction cost of Vanilla and \sys on Pony and Psychology using our Python DiskANN implementation with stored Diver embeddings.
We use the default $R=16$, $B=2$, $M=24$, $K=4$, and $g_{\min}=12$, with the full pruning budget $\beta=1$.
Each run uses eight CPU cores and one NVIDIA H200, with Qwen3-32B for semantic selection.
As shown in~\Cref{tab:pruning-cost}, \sys constructs the Pony and Psychology indices in 109.46 and 691.40 minutes, compared with 1.49 and 11.84 minutes for Vanilla.
More than 96\% of the \sys construction time is spent waiting for LLM responses.
At the full pruning budget, \sys makes one LLM call per node on average in both measured runs, with average graph-construction times of 0.83 and 0.79 seconds per node on Pony and Psychology, respectively.
The additional cost is entirely offline: once constructed, the graph can be reused across queries without further calls to the semantic-selection model, while the query-time graph-search procedure remains unchanged.

\begin{table*}[t]
\centering
\small
\setlength{\tabcolsep}{4pt}
\renewcommand{\arraystretch}{1.04}
\caption{
Distance-computation-budget sweeps on BRIGHT and M-BEIR using Diver and CLIP embeddings, respectively.
Results are averaged equally across datasets and tasks.
Reranking uses the top 20 retrieved candidates at each budget.
\textcolor{gainGreen}{Green} subscripts denote relative improvements over Vanilla, computed before rounding.
\textcolor{exactGray}{Gray} values denote average exact-$k$NN NDCG@10 / Recall@10 without reranking.
}
\label{tab:computation-budget}
\scalebox{0.84}{
\begin{tabular}{lccccccccc}
\toprule
\multirow{3}{*}[-1ex]{
\makecell[c]{
Benchmark\\[-1pt]
\footnotesize\textcolor{exactGray}{(Exact $k$NN ref.)}
}
}
& \multirow{3}{*}[-1ex]{\makecell[c]{Computation\\budget}}
& \multicolumn{4}{c}{Greedy search}
& \multicolumn{4}{c}{Rerank} \\
\cmidrule(lr){3-6}\cmidrule(lr){7-10}
& & \multicolumn{2}{c}{NDCG@10}
& \multicolumn{2}{c}{Recall@10}
& \multicolumn{2}{c}{NDCG@10}
& \multicolumn{2}{c}{Recall@10} \\
\cmidrule(lr){3-4}\cmidrule(lr){5-6}
\cmidrule(lr){7-8}\cmidrule(lr){9-10}
& & Vanilla & \sys
& Vanilla & \sys
& Vanilla & \sys
& Vanilla & \sys \\
\midrule
\multirow{7}{*}{
\makecell[l]{
BRIGHT\\[-1pt]
\footnotesize\textcolor{exactGray}{(28.50\,/\,34.18)}
}
}
& 100
& 3.96 & \lgpgain{5.02}{26.9}
& 2.96 & \lgpgain{4.15}{40.0}
& 4.28 & \lgpgain{5.40}{26.1}
& 3.07 & \lgpgain{4.12}{34.3} \\
& 200
& 10.50 & \lgpgain{13.98}{33.2}
& 11.62 & \lgpgain{15.60}{34.3}
& 11.67 & \lgpgain{15.41}{32.0}
& 12.32 & \lgpgain{16.40}{33.1} \\
& 300
& 16.69 & \lgpgain{19.80}{18.6}
& 19.18 & \lgpgain{23.19}{20.9}
& 18.70 & \lgpgain{22.46}{20.1}
& 20.35 & \lgpgain{24.61}{21.0} \\
& 500
& 22.92 & \lgpgain{24.57}{7.2}
& 27.14 & \lgpgain{28.83}{6.2}
& 26.04 & \lgpgain{27.79}{6.7}
& 28.70 & \lgpgain{30.82}{7.4} \\
& 800
& 25.89 & \lgpgain{26.61}{2.8}
& 30.66 & \lgpgain{31.37}{2.3}
& 29.00 & \lgpgain{29.92}{3.2}
& 32.59 & \lgpgain{33.54}{2.9} \\
& 1600
& 27.11 & \lgpgain{27.27}{0.6}
& 32.27 & \lgpgain{32.32}{0.2}
& 30.83 & \lgpgain{30.95}{0.4}
& 34.63 & \lgpgain{34.71}{0.2} \\
& 3200
& 27.65 & \lgpgain{27.74}{0.3}
& 32.90 & \lgpgain{32.96}{0.2}
& 31.26 & \lgpgain{31.28}{0.1}
& 35.15 & \lgpgain{35.18}{0.1} \\
\midrule
\multirow{7}{*}{\makecell[l]{M-BEIR\\[-1pt]\footnotesize\textcolor{exactGray}{(16.16\,/\,22.87)}}}
& 100
& 2.86 & \lgpgain{2.97}{3.8}
& 3.88 & \lgpgain{4.14}{6.8}
& 3.53 & \lgpgain{3.69}{4.4}
& 4.21 & \lgpgain{4.59}{8.9} \\
& 200
& 7.60 & \lgpgain{7.71}{1.4}
& 10.89 & \lgpgain{10.97}{0.8}
& 10.13 & \lgpgain{10.35}{2.1}
& 12.47 & \lgpgain{12.56}{0.7} \\
& 300
& 9.56 & \lgpgain{10.04}{5.0}
& 13.61 & \lgpgain{14.42}{5.9}
& 12.61 & \lgpgain{13.49}{7.0}
& 15.32 & \lgpgain{16.40}{7.1} \\
& 500
& 11.60 & \lgpgain{12.57}{8.4}
& 16.52 & \lgpgain{17.69}{7.1}
& 15.33 & \lgpgain{17.01}{11.0}
& 18.52 & \lgpgain{20.41}{10.2} \\
& 800
& 13.28 & \lgpgain{13.70}{3.2}
& 18.50 & \lgpgain{19.25}{4.0}
& 17.79 & \lgpgain{18.58}{4.5}
& 21.46 & \lgpgain{22.16}{3.2} \\
& 1600
& 14.44 & \lgpgain{14.98}{3.7}
& 20.10 & \lgpgain{20.85}{3.7}
& 19.53 & \lgpgain{20.26}{3.7}
& 23.51 & \lgpgain{24.12}{2.6} \\
& 3200
& 15.27 & \lgpgain{15.51}{1.6}
& 21.17 & \lgpgain{21.48}{1.5}
& 20.59 & \lgpgain{20.80}{1.0}
& 24.59 & \lgpgain{24.75}{0.7} \\
\bottomrule
\end{tabular}
}
\vspace{-14pt}
\end{table*}

\subsection{Evaluation under Fixed Distance-Computation Budgets}\label{app:comp}
Search width is the standard query-time control parameter for graph-based ANN indices such as DiskANN and HNSW.
By controlling the size of the candidate set explored during graph search, it directly regulates search effort: with graph degree bounded by $R$, the number of distance computations scales roughly as $O(\textsc{search width}\times R)$.
We therefore use search width as the primary query-time budget parameter throughout our main evaluation.
Because the exact number of distance computations at a fixed width may vary with the graph topology, we additionally provide a complementary comparison of Vanilla and \sys under the same explicit query--document distance-computation budgets.

Specifically, we vary the distance-computation budget over
$\{100,200,300,500,800,1600,3200\}$ and evaluate Vanilla and \sys on BRIGHT and M-BEIR using Diver and CLIP embeddings, respectively, with DiskANN.
We report both greedy-search performance and performance after reranking the top 20 retrieved candidates.
As shown in~\Cref{tab:computation-budget}, \sys consistently outperforms Vanilla at every tested computation budget under both greedy search and reranking on both benchmarks.
The gains are largest at smaller and intermediate budgets.
For example, on BRIGHT at a budget of 200 distance computations, \sys improves greedy-search NDCG@10 and Recall@10 by \textbf{33.2\%} and \textbf{34.3\%}, while reranked NDCG@10 and Recall@10 improve by \textbf{32.0\%} and \textbf{33.1\%}, respectively.
Moreover, on Economics from BRIGHT and OVEN Task~8 from M-BEIR, \sys outperforms Vanilla in NDCG@10 under both greedy search and reranking with only half the distance-computation budget (800 vs.\ 1,600 per query), corresponding to 50\% fewer computations.
As the computation budget increases and both methods approach exact retrieval, the gap naturally narrows.
These results show that \sys improves search efficiency by achieving higher retrieval quality under the same computation budget and comparable quality with substantially fewer distance evaluations.

\section{Prompts}\label{app:prompts}
\subsection{Prompt used in \textsc{LLMSelect}}\label{app:shortcut-prompt}
The \textsc{LLMSelect} operation in our algorithm uses the prompt below.


\begin{tcolorbox}[title={System Prompt}, left=1mm, right=1mm, top=1mm, bottom=1mm, boxrule=0.9pt, breakable]
\begin{Verbatim}[breaklines=true]
You are editing a theorem retrieval graph at indexing time.
There is no real user query available, and you must not assume a particular test query.
Instead, privately simulate plausible latent user queries / problem intents for which the ANCHOR might be retrieved as a nearby theorem.

Your goal is to improve query-time routing from the ANCHOR to useful answers.
First, if helpful, optionally write a short latent query line beginning with QUERY: ; then select at most the best candidates for shortcut edges.

For each CANDIDATE, ask: would a user query that lands near the ANCHOR also reasonably need this CANDIDATE as the target theorem, a prerequisite theorem, a bridge theorem, a generalization/special case, or a theorem used in the same proof/problem-solving route?

Select a CANDIDATE only if adding an edge ANCHOR -> CANDIDATE would help query-time graph search route from the ANCHOR's local neighborhood toward useful answers that are not already covered by the ALREADY-SELECTED LOCAL NEIGHBORS.

Do NOT select based on wording similarity alone.
Do NOT select broad but unrelated facts.
If no candidate has a clear latent-query/use-case connection, return NONE.
Return at most {B} candidate numbers, best first, or NONE. Examples: [3] [1] or NONE
\end{Verbatim}
\end{tcolorbox}

\begin{tcolorbox}[title={User Prompt}, left=1mm, right=1mm, top=1mm, bottom=1mm, boxrule=0.9pt, breakable]
\begin{Verbatim}[breaklines=true]
ANCHOR:
{anchor}

ALREADY-SELECTED LOCAL NEIGHBORS:
{keep_block}

CANDIDATE SHORTCUTS:
{cand_block}

Privately imagine the latent queries/use-cases, then choose which candidates should be connected to the ANCHOR for query-time routing. Return only candidate numbers or NONE.
If you include a latent query, start it with QUERY: on its own line.
\end{Verbatim}
\end{tcolorbox}

\subsection{Text Reranking Prompt}\label{app:rerank-prompt}

\begin{tcolorbox}[title={Text Reranking: System Prompt}, left=1mm, right=1mm, top=1mm, bottom=1mm, boxrule=0.9pt, breakable]
\begin{Verbatim}[breaklines=true,fontsize=\small]
You are an intelligent assistant that can rank answers based on their relevancy to the query.
I will provide you with {N} passages, each indicated by number identifier [].
Rank the answers based on their relevance to query: {QUERY}.
\end{Verbatim}
\end{tcolorbox}

\begin{tcolorbox}[title={Text Reranking: User Prompt}, left=1mm, right=1mm, top=1mm, bottom=1mm, boxrule=0.9pt, breakable]
\begin{Verbatim}[breaklines=true,fontsize=\small]
[1] {passage_1}

...

[{N}] {passage_N}

Query: {QUERY}.
Rank the {N} passages above based on their relevance to the query.
The passages should be listed in descending order using identifiers.
The most relevant passages should be listed first.
The output format should be like [1] > [2] ... > [{N}].
Only response the ranking results, do not say any word or explain.
\end{Verbatim}
\end{tcolorbox}

\subsection{Multimodal Reranking Prompt}\label{app:mm-rerank-prompt}
\begin{tcolorbox}[title={Multimodal Reranking: System Prompt}, left=1mm, right=1mm, top=1mm, bottom=1mm, boxrule=0.9pt, breakable]
\begin{Verbatim}[breaklines=true,fontsize=\small]
You rank retrieval candidates for multimodal retrieval.
Given one QUERY and several CANDIDATES, rank the candidates from best to worst match.
Use all available evidence, including image and text when present.
Return only an ordering of candidate indices like [2] [1] [3].
\end{Verbatim}
\end{tcolorbox}

\begin{tcolorbox}[title={Multimodal Reranking: User Content}, left=1mm, right=1mm, top=1mm, bottom=1mm, boxrule=0.9pt, breakable]
\begin{Verbatim}[breaklines=true,fontsize=\small]
QUERY:
Task instruction:
{task_instruction}
Query text:
{query_text}
<query image>

CANDIDATES:
[1]
Candidate 1 text:
{candidate_1_text}
<candidate 1 image>
...
[{N}]
Candidate {N} text:
{candidate_N_text}
<candidate N image>

Rank the CANDIDATES from best to worst match for the QUERY.
Use image and text jointly when available.
Return only candidate indices like [2] [1] [3].
\end{Verbatim}
\end{tcolorbox}

\section{Algorithms of DiskANN and HNSW}
\label{app:original}

We provide the detailed algorithms of DiskANN in~\Cref{alg:diskann_original,alg:greedy-search,alg:robust-pruning} and HNSW in~\Cref{alg:hnsw_original,alg:hnsw_search_layer,alg:hnsw_select_neighbors}.
For HNSW, $N_\ell(i)$ denotes the outgoing neighborhood of $v_i$ at layer $\ell$, and the unlayered notation $N(i)$ used in our base-layer pruning refers to $N_0(i)$.
We denote the construction search width by $L_{\mathrm{build}}$ and use a generic width argument $L$ in the search routines below.
DiskANN construction uses $L=L_{\mathrm{build}}$.
HNSW insertion uses $L=1$ when descending through layers above the new node's maximum level, and $L=L_{\mathrm{build}}$ when searching layers in which the node is inserted.
At query time, DiskANN uses $L=L_{\mathrm{search}}$.
HNSW uses $L=1$ for upper-layer descent and $L=ef_{\mathrm{search}}$ for base-layer search.

\begin{algorithm}[ht]
\caption{DiskANN Indexing}
\label{alg:diskann_original}
\begin{algorithmic}[1]

\STATE \textbf{Input} Document embeddings $\{\bm e(d_i)\}_{i=1}^{|V|}$, degree budget $R$, construction search width $L_{\mathrm{build}}$, pruning parameter $\alpha$
\STATE \textbf{Result} Graph index $G=(V,E)$ with local neighborhoods $\{N(i)\}_{i=1}^{|V|}$

\STATE Initialize $G$ as a random $R$-regular graph over $V=\{v_1,\dots,v_{|V|}\}$
\STATE Let $s$ be the vertex whose embedding is closest to the centroid of $\{\bm e(d_i)\}_{i=1}^{|V|}$

\FOR{$t=1$ to $2$}
    \STATE Sample a random permutation $\sigma$ of $\{1,\dots,|V|\}$
    \FOR{$k=1$ to $|V|$}
        \STATE $i \leftarrow \sigma(k)$
        \STATE $U_i \leftarrow \textsc{GreedySearch}(s,v_i,L_{\mathrm{build}})$
        \STATE $N(i) \leftarrow \textsc{RobustPruning}(v_i,U_i,\alpha,R)$
        \FOR{each $v_j \in N(i)$}
            \STATE $N(j) \leftarrow N(j) \cup \{v_i\}$
            \IF{$|N(j)| > R$}
                \STATE $N(j) \leftarrow \textsc{RobustPruning}(v_j,N(j),\alpha,R)$
            \ENDIF
        \ENDFOR
    \ENDFOR
\ENDFOR

\end{algorithmic}
\end{algorithm}

\begin{algorithm}[ht]
\caption{\textsc{GreedySearch} in DiskANN}
\label{alg:greedy-search}
\begin{algorithmic}[1]

\STATE \textbf{Input} Start point $s$, query point $q$, search width $L$
\STATE \textbf{Result} Visited vertex list $U$

\STATE $A \gets \{s\}$, \quad $U \gets \varnothing$

\WHILE{$A \setminus U \neq \varnothing$}
    \STATE $v_j \gets \arg\min_{x \in A \setminus U} D(x,q)$
    \STATE $A \gets A \cup N(j)$
    \STATE $U \gets U \cup \{v_j\}$

    \IF{$|A| > L$}
        \STATE $A \gets$ the $L$ vertices in $A$ closest to $q$
    \ENDIF
\ENDWHILE

\STATE Sort $U$ in increasing distance from $q$
\RETURN $U$

\end{algorithmic}
\end{algorithm}

\begin{algorithm}[!th]
\caption{\textsc{RobustPruning} in DiskANN}
\label{alg:robust-pruning}
\begin{algorithmic}[1]

\STATE \textbf{Input} Vertex $v_i$, candidate neighbor set $U$, pruning parameter $\alpha$, degree limit $R$
\STATE \textbf{Result} Updated $N(i)$, the set of out-neighbors of $v_i$

\STATE $U \gets \left(U \cup N(i)\right)\setminus\{v_i\}$
\STATE $N(i) \gets \varnothing$

\WHILE{$U \neq \varnothing$ \AND $|N(i)| < R$}
    \STATE $v_j \gets \arg\min_{x \in U} D(x,v_i)$
    \STATE $N(i) \gets N(i) \cup \{v_j\}$
    \STATE $U \gets U \setminus \{v_j\}$
    \STATE $U \gets \{v_k \in U : \alpha D(v_j,v_k) > D(v_i,v_k)\}$
\ENDWHILE

\RETURN $N(i)$

\end{algorithmic}
\end{algorithm}

\begin{algorithm}[ht]
\caption{HNSW Insertion}
\label{alg:hnsw_original}
\begin{algorithmic}[1]

\STATE \textbf{Input} Vertex $v_i$, current entry point $s$, neighbor budget $R$, base-layer degree cap $R_0$, construction search width $L_{\mathrm{build}}$, level-scale parameter $m_\ell$, flags \textsc{ExtendCandidates} and \textsc{KeepPruned}
\STATE \textbf{Result} Updated HNSW graph after inserting $v_i$

\STATE Sample the maximum level $\ell_i$ of $v_i$ according to the level distribution parameterized by $m_\ell$
\STATE Let $\ell_{\max}$ be the level of the current entry point
\STATE $S_{\mathrm{entry}} \leftarrow \{s\}$

\FOR{$\ell=\ell_{\max},\ell_{\max}-1,\dots,\ell_i+1$}
    \STATE $W \leftarrow \textsc{Search-Layer}(v_i,S_{\mathrm{entry}},1,\ell)$
    \STATE $S_{\mathrm{entry}} \leftarrow \{\arg\min_{x \in W} D(x,v_i)\}$
\ENDFOR

\FOR{$\ell=\min(\ell_{\max},\ell_i),\min(\ell_{\max},\ell_i)-1,\dots,0$}
    \STATE $W \leftarrow \textsc{Search-Layer}(v_i,S_{\mathrm{entry}},L_{\mathrm{build}},\ell)$
    \STATE $N_\ell(i) \leftarrow \textsc{Select-Neighbors}(v_i,W,R,\ell,\textsc{ExtendCandidates},\textsc{KeepPruned})$
    \STATE Add bidirectional connections between $v_i$ and all nodes in $N_\ell(i)$ at layer $\ell$

    \FOR{each $v_j \in N_\ell(i)$}
        \STATE Let $N_\ell(j)$ be the current neighborhood of $v_j$ at layer $\ell$

        \IF{$\ell=0$ \AND $|N_\ell(j)|>R_0$}
            \STATE $N_\ell(j) \leftarrow \textsc{Select-Neighbors}(v_j,N_\ell(j),R_0,\ell,\textsc{ExtendCandidates},\textsc{KeepPruned})$
        \ELSIF{$\ell>0$ \AND $|N_\ell(j)|>R$}
            \STATE $N_\ell(j) \leftarrow \textsc{Select-Neighbors}(v_j,N_\ell(j),R,\ell,\textsc{ExtendCandidates},\textsc{KeepPruned})$
        \ENDIF
    \ENDFOR

    \STATE $S_{\mathrm{entry}} \leftarrow W$
\ENDFOR

\IF{$\ell_i>\ell_{\max}$}
    \STATE Set the entry point of the graph to $v_i$
\ENDIF

\end{algorithmic}
\end{algorithm}

\begin{algorithm}[th]
\caption{\textsc{Search-Layer} in HNSW}
\label{alg:hnsw_search_layer}
\begin{algorithmic}[1]

\STATE \textbf{Input} Query $q$, entry-point set $S_{\mathrm{entry}}$, search width $L$, layer $\ell$
\STATE \textbf{Result} Candidate set $W$

\STATE $V_{\mathrm{seen}} \leftarrow S_{\mathrm{entry}}$, $\mathcal{C} \leftarrow S_{\mathrm{entry}}$, $W \leftarrow S_{\mathrm{entry}}$

\WHILE{$\mathcal{C} \neq \varnothing$}
    \STATE $v_j \leftarrow \arg\min_{x \in \mathcal{C}} D(x,q)$
    \STATE $\mathcal{C} \leftarrow \mathcal{C}\setminus\{v_j\}$
    \STATE $v_f \leftarrow \arg\max_{x \in W} D(x,q)$

    \IF{$D(v_j,q)>D(v_f,q)$}
        \STATE \textbf{break}
    \ENDIF

    \FOR{each $v_k \in N_\ell(j)$}
        \IF{$v_k \notin V_{\mathrm{seen}}$}
            \STATE $V_{\mathrm{seen}} \leftarrow V_{\mathrm{seen}}\cup\{v_k\}$
            \STATE $v_f \leftarrow \arg\max_{x \in W} D(x,q)$

            \IF{$|W|<L$ \OR $D(v_k,q)<D(v_f,q)$}
                \STATE $\mathcal{C} \leftarrow \mathcal{C}\cup\{v_k\}$
                \STATE $W \leftarrow W\cup\{v_k\}$

                \IF{$|W|>L$}
                    \STATE $v_f \leftarrow \arg\max_{x \in W} D(x,q)$
                    \STATE $W \leftarrow W\setminus\{v_f\}$
                \ENDIF
            \ENDIF
        \ENDIF
    \ENDFOR
\ENDWHILE

\RETURN $W$

\end{algorithmic}
\end{algorithm}

\begin{algorithm}[th]
\caption{\textsc{Select-Neighbors} in HNSW}
\label{alg:hnsw_select_neighbors}
\begin{algorithmic}[1]

\STATE \textbf{Input}
Vertex $v_i$, candidate set $\mathcal{C}$, neighbor budget $R$,
layer $\ell$, flags \textsc{ExtendCandidates} and \textsc{KeepPruned}

\STATE \textbf{Result}
Selected neighbor set $S$

\STATE $S \leftarrow \varnothing$
\STATE $W \leftarrow \mathcal{C}$

\IF{\textsc{ExtendCandidates}}
    \FOR{each $v_j\in\mathcal{C}$}
        \FOR{each $v_k\in N_\ell(j)$}
            \IF{$v_k\notin W$}
                \STATE $W\leftarrow W\cup\{v_k\}$
            \ENDIF
        \ENDFOR
    \ENDFOR
\ENDIF

\STATE $W_{\mathrm{discard}}\leftarrow\varnothing$

\WHILE{$W\neq\varnothing$ \AND $|S|<R$}
    \STATE $v_j\leftarrow\arg\min_{x\in W}D(x,v_i)$
    \STATE $W\leftarrow W\setminus\{v_j\}$

    \IF{$D(v_j,v_i)\leq D(v_j,v_k),\ \forall v_k\in S$}
        \STATE $S\leftarrow S\cup\{v_j\}$
    \ELSE
        \STATE $W_{\mathrm{discard}}
        \leftarrow W_{\mathrm{discard}}\cup\{v_j\}$
    \ENDIF
\ENDWHILE

\IF{\textsc{KeepPruned}}
    \WHILE{$W_{\mathrm{discard}}\neq\varnothing$ \AND $|S|<R$}
        \STATE
        $v_j\leftarrow
        \arg\min_{x\in W_{\mathrm{discard}}}D(x,v_i)$
        \STATE
        $W_{\mathrm{discard}}
        \leftarrow W_{\mathrm{discard}}\setminus\{v_j\}$
        \STATE $S\leftarrow S\cup\{v_j\}$
    \ENDWHILE
\ENDIF

\RETURN $S$

\end{algorithmic}
\end{algorithm}

\FloatBarrier

\FloatBarrier

\begin{algorithm}[!t]
\small
\caption{LLM-Guided Graph Pruning (\sys)}
\label{alg:llm_pruning_detailed}
\begin{algorithmic}[1]

\STATE \textbf{Input} ANN graph $G$, current node $v_i$, neighborhood $N(i)$, and index-specific candidate set $U_i$
\STATE \textbf{Parameters} Degree cap $R$, shortlist budget $M$, per-node edit budget $B$, neighborhood context size $K$, and threshold $g_{\min}$

\STATE {\ttfamily \textcolor{red}{\(\triangleright\) {/* Form a local replacement set */}}}

\STATE
$T_i \leftarrow
\left(
    \bigcup_{v_j\in N(i)} N(j)
\right)
\setminus \{v_i\}$

\STATE
$U_i \leftarrow
\left(U_i \cup T_i\right)
\setminus
\left(N(i)\cup\{v_i\}\right)$

\STATE {\ttfamily \textcolor{red}{\(\triangleright\) {/* Shortlist local replacement candidates without LLM calls */}}}

\FOR{each $v_c\in U_i$}

    \STATE
    $g_i(c)\leftarrow
    |N(c)\setminus T_i|$

    \STATE
    $\nu_i(c)\leftarrow
    \min_{v_j\in N(i)} D(v_c,v_j)$

    \STATE
    $\delta_i(c)\leftarrow
    D(v_i,v_c)$

\ENDFOR

\STATE
$L_g\leftarrow
\textsc{Sort}_{\downarrow}(U_i;g_i)$

\STATE
$L_{\nu}\leftarrow
\textsc{Sort}_{\downarrow}(U_i;\nu_i)$

\STATE
$L_{\delta}\leftarrow
\textsc{Sort}_{\downarrow}(U_i;\delta_i)$

\STATE
$m_{\mathrm{src}}\leftarrow \lfloor M/3\rfloor$

\STATE
$U_i\leftarrow$
the first $M$ distinct candidates in
$L_{\delta}[1\!:\!m_{\mathrm{src}}]
\,\Vert\,
L_g[1\!:\!m_{\mathrm{src}}]
\,\Vert\,
L_{\nu}$

\STATE {\ttfamily \textcolor{red}{\(\triangleright\) {/* Select semantically useful candidates with an LLM */}}}

\STATE
$\K_i\leftarrow$ the $\min(K, |N(i)|)$ nodes in $N(i)$ closest to $v_i$
\STATE  
$\mathcal{A}_i\leftarrow
\textsc{LLMSelect}(v_i,\K_i,U_i,B)$








\STATE {\ttfamily \textcolor{red}{\(\triangleright\) {/* Local edge pruning and replacement */}}}

\STATE
$N'(i)\leftarrow N(i)$

\FOR{each $v_c\in\mathcal{A}_i$ in selection order}

    \IF{$v_c\in N'(i)$ \OR $g_i(c)<g_{\min}$}
        \STATE \textbf{continue}
    \ENDIF


    \IF{$|N'(i)|<R$}

        \STATE
        $N'(i)\leftarrow N'(i)\cup\{v_c\}$

        \STATE \textbf{continue}

    \ENDIF

    \IF{$N'(i)\cap N(i)=\varnothing$}
        \STATE \textbf{continue}
    \ENDIF

    \FOR{each original neighbor $v_j\in N'(i)\cap N(i)$}

        \STATE
        $\ell_i(j)\leftarrow
        \left|
            N(j)\setminus
            \left(
                \left(
                    \bigcup_{v_k\in N'(i)\setminus\{v_j\}} N(k)
                \right)
                \setminus\{v_i\}
            \right)
        \right|$

        \STATE
        $\nu_i^{-j}(j)\leftarrow
        \min_{v_k\in N'(i)\setminus\{v_j\}}
        D(v_j,v_k)$

    \ENDFOR

    \STATE
    $v_w\leftarrow
    \arg\min^{\mathrm{lex}}_{v_j\in N'(i)\cap N(i)}
    \left(
        \ell_i(j),
        \nu_i^{-j}(j)
    \right)$

    \STATE
    $N'(i)\leftarrow
    \left(N'(i)\setminus\{v_w\}\right)
    \cup\{v_c\}$

\ENDFOR

\STATE \textbf{return} $N'(i)$

\end{algorithmic}
\end{algorithm}


\section{Detailed \sys Algorithms and Index-Specific Instantiations}\label{app:algos}

\Cref{alg:llm_pruning_detailed} provides the detailed algorithm for \sys.
We first describe its candidate construction, semantic selection, and edge-replacement steps, and then show how the same local update is instantiated in DiskANN and HNSW.

\noindent\textbf{Efficient Local Candidate Set Construction.}
For each node $v_i$ in the given ANN graph $G$, \sys forms a local set of possible replacement neighbors $\C_i$ rather than considering every node in the graph. 
Since most graph-based ANN methods already produce a candidate set when selecting or updating the outgoing neighborhood $N(i)$ of $v_i$, \sys reuses this set, denoted by $U_i$, as its initial replacement set and augments it with the nodes exposed through $N(i)$ (lines 3-5 in \Cref{alg:llm_pruning_detailed}):
\begin{equation}
    T_i:=\left(\bigcup_{v_j\in N(i)}N(j)\right)\setminus\{v_i\},
    \qquad
    U_i:=\left(U_i\cup T_i\right)\setminus\left(N(i)\cup\{v_i\}\right).
\end{equation}
Here, $T_i$ contains the nodes reachable from $v_i$ in two directed hops through its current neighbors, excluding $v_i$ itself.
After removing existing neighbors, the updated $U_i$ therefore contains local non-neighbor alternatives drawn from both the index-specific candidate set and the directed two-hop neighborhood.

To bound the cost of LLM-guided selection and avoid excessively long contexts, \sys further reduces $U_i$ to at most $M$ candidates. (lines 6--15 in \Cref{alg:llm_pruning_general}). 
For each candidate node $v_c\in U_i$, \sys considers three complementary ranking signals:
\begin{equation}
    \delta_i(c) := D(v_i,v_c), \qquad
    g_i(c) := |N(c)\setminus T_i|, \qquad
    \nu_i(c) := \min_{v_j\in N(i)} D(v_c,v_j).
    \label{eq:shortcut-proxies}
\end{equation}
Here, $\delta_i(c)$ measures proximity to $v_i$, $g_i(c)$ measures the additional two-hop coverage introduced by $v_c$, and $\nu_i(c)$ is its distance to the closest current neighbor, with larger values indicating less geometric redundancy.
Let $L_g$, $L_{\nu}$, and $L_{\delta}$ denote the rankings of $U_i$ by decreasing $g_i(c)$, decreasing $\nu_i(c)$, and increasing $\delta_i(c)$, respectively. 
\sys forms $\mathcal{P}_i$ by taking up to $M/3$ candidates from $L_{\delta}$, up to $\lfloor M/3\rfloor$ from $L_g$, and filling the remaining positions from $L_{\nu}$.

\noindent\textbf{LLM-Guided Semantic Selection.}
Given the compact candidate shortlist, \sys provides the LLM with the document represented by $v_i$, a compact neighborhood context $\K_i$ consisting of the $\min(K,|N(i)|)$ neighbors in $N(i)$ closest to $v_i$ in embedding space, and the candidate documents in $U_i$.
The LLM is then asked to select at most $B$ candidates that extend rather than repeat the information already represented in $\K_i$, for example by providing missing supporting context, connecting $v_i$ to a complementary concept, or extending its content toward a useful specialization or next step.
The selected candidates form a proposal replacement set $\mathcal{A}_i\subseteq U_i$ (lines~16--18 in~\Cref{alg:llm_pruning_detailed}).

\begin{algorithm}[t]
\caption{
DiskANN construction with \sys.
\textcolor{lgpcolor}{Colored} lines indicate operations introduced by \sys.
}
\label{alg:diskann_lgp}
\begin{algorithmic}[1]

\STATE \textbf{Input}
Document embeddings $\{\bm e(d_i)\}_{i=1}^{|V|}$,
degree budget $R$,
construction search width $L_{\mathrm{build}}$,
pruning parameter $\alpha$,
and \sys parameters $(B,M,K,g_{\min})$

\STATE \textbf{Result}
Graph index $G=(V,E)$ with refined neighborhoods
$\{N(i)\}_{i=1}^{|V|}$

\STATE Initialize $G$ as a random $R$-regular graph and choose the centroid-nearest entry point $s$

\FOR{$t=1$ to $2$}
    \STATE Sample a random permutation $\sigma$ of $\{1,\dots,|V|\}$

    \FOR{$k=1$ to $|V|$}
        \STATE $i\leftarrow\sigma(k)$
        \STATE $U_i\leftarrow
        \textsc{GreedySearch}(s,v_i,L_{\mathrm{build}})$

        \STATE $N(i)\leftarrow
        \textsc{RobustPruning}(v_i,U_i,\alpha,R)$

        \STATE \textcolor{lgpcolor}{$N_{\mathrm{pre}}(i)\leftarrow N(i)$}

        \IF{$t=2$}
            \STATE \textcolor{lgpcolor}{
            $N(i)\leftarrow
            \textsc{LGP}
            (G,v_i,N(i),U_i,R,M,B,K,g_{\min})$
            }
        \ENDIF

        \FOR{each $v_j\in N(i)$}
            \STATE Insert $v_i$ into $N(j)$ if absent

            \IF{$|N(j)|>R$}

                \IF{$t=2$ \AND $v_j\in N(i)\setminus N_{\mathrm{pre}}(i)$}
                    \STATE \textcolor{lgpcolor}{
                    $N(j)\leftarrow
                    \textsc{RobustPruning-Protected}
                    (v_j,N(j),\alpha,R,v_i)$
                    }
                \ELSE
                    \STATE
                    $N(j)\leftarrow
                    \textsc{RobustPruning}(v_j,N(j),\alpha,R)$
                \ENDIF

            \ENDIF
        \ENDFOR
    \ENDFOR
\ENDFOR

\RETURN $G$

\end{algorithmic}
\end{algorithm}

\begin{algorithm}[!t]
\caption{
Layer-0 HNSW pruning with \sys.
\textcolor{lgpcolor}{Colored} lines indicate operations introduced by \sys.
}
\label{alg:hnsw_lgp}
\begin{algorithmic}[1]

\STATE \textbf{Input}
Completed Vanilla HNSW graph $G$,
construction search width $L_{\mathrm{build}}$,
and \sys parameters $(B,M,K,g_{\min})$

\STATE \textbf{Result}
Refined HNSW graph $G$ with unchanged upper layers

\STATE \lgpadd{Freeze all upper layers, node levels, and the entry point}
\STATE \lgpadd{Sample a random permutation $\sigma$ of $\{1,\dots,|V|\}$}

\FOR{$k=1$ to $|V|$}

    \STATE \lgpadd{$i\leftarrow\sigma(k)$}
    \STATE \lgpadd{$r_i\leftarrow |N_0(i)|$}

    \IF{$r_i=0$}
        \STATE \textbf{continue}
    \ENDIF

    \STATE \lgpadd{
    $U_i\leftarrow$ the base-layer candidates reached by hierarchical
    search for $v_i$ using width $L_{\mathrm{build}}$
    }

    \STATE \lgpadd{$N_{\mathrm{pre}}(i)\leftarrow N_0(i)$}

    \STATE \lgpadd{
    $N_0(i)\leftarrow
    \textsc{LGP-Local}
    (G,v_i,N_0(i),U_i,r_i,M,B,K,g_{\min})$
    }

    \STATE \lgpadd{Assert $|N_0(i)|=r_i$}

    \FOR{each $v_c\in N_0(i)\setminus N_{\mathrm{pre}}(i)$
         such that $v_i\notin N_0(c)$}

        \STATE \lgpadd{
        $H_c\leftarrow
        \textsc{Select-Neighbors}
        (v_c,N_0(c)\cup\{v_i\},|N_0(c)|,0,
        \textsc{false},\textsc{false})$
        }

        \IF{$v_i\in H_c$}

            \IF{$N_0(c)\setminus H_c\neq\varnothing$}

                \STATE \lgpadd{
                $v_z\leftarrow
                \arg\max_{v_j\in N_0(c)\setminus H_c}
                D(v_c,v_j)$
                }

                \STATE \lgpadd{
                $N_0(c)\leftarrow
                \left(N_0(c)\setminus\{v_z\}\right)\cup\{v_i\}$
                }

            \ENDIF

        \ENDIF

    \ENDFOR
\ENDFOR

\STATE \lgpadd{
Verify that all base-layer degrees and all upper layers are unchanged
}

\RETURN $G$

\end{algorithmic}
\end{algorithm}

\noindent\textbf{Local Edge Replacement.}
Before incorporating an LLM-selected candidate, \sys applies a lightweight structural filter with threshold $g_{\min}$, retaining only candidates that provide sufficient additional two-hop coverage.
Let $N'(i)$ denote the current working neighborhood of $v_i$. 
For each filtered candidate $v_c\in \mathcal{A}_i$, \sys adds it directly when $|N'(i)|<R$, and otherwise replaces a low-value original neighbor $v_w$ (lines~19--33 in~\Cref{alg:llm_pruning_detailed}).

For each original neighbor $v_j\in N'(i)\cap N(i)$, \sys measures its unique two-hop contribution and its redundancy with the remaining neighborhood:
\begin{equation}
    \ell_i(j) := \left|N(j)\setminus \left( \left(\bigcup_{v_k\in N'(i)\setminus\{v_j\}} N(k)\right) \setminus\{v_i\} \right) \right|, \qquad
    \nu_i^{-j}(j) := \min_{v_k\in N'(i)\setminus\{v_j\}} D(v_j,v_k).
\end{equation}
Here, $\ell_i(j)$ measures the unique two-hop contribution of $v_j$ relative to the remaining neighborhood, while $\nu_i^{-j}(j)$ measures its distance to the closest remaining neighbor.
\sys selects the neighbor $v_w$ to remove by lexicographically minimizing $\bigl(\ell_i(j), \nu_i^{-j}(j)\bigr)$.
This identifies neighbors with little unique local coverage and high redundancy as the lowest-value candidates for replacement.

\subsection{Instantiation for Representative ANN Indices}
\label{app:ins}

\Cref{alg:llm_pruning_detailed} can be inserted into different graph-based ANN indices without changing their query-time search procedures.
\Cref{alg:diskann_lgp,alg:hnsw_lgp} give the concrete instantiations for DiskANN and HNSW, respectively.

\noindent\textbf{DiskANN.}
As shown in~\Cref{alg:diskann_lgp}, \sys is applied immediately after \textsc{RobustPruning} during the second DiskANN construction pass.
The vertices visited by \textsc{GreedySearch} are reused as the candidate set $U_i$, after which \textsc{LGP} refines $N(i)$.
Newly introduced edges follow DiskANN's usual reciprocal-insertion step. 
If a reciprocal insertion exceeds the degree bound, the newly inserted reciprocal edge is protected only during that immediate pruning operation.
The protected pruning operation retains this edge while pruning the other candidates to satisfy the degree bound $R$. 
This protection does not persist beyond that pruning operation.

\noindent\textbf{HNSW.}
As shown in~\Cref{alg:hnsw_lgp}, \sys is applied as an offline pruning of layer~0 after the vanilla HNSW index has been constructed.
For each node $v_i$, hierarchical search is reused to obtain the candidate set $U_i$, and \textsc{LGP} refines only its base-layer neighborhood $N_0(i)$ while preserving its original degree.
For each newly introduced base-layer edge, HNSW's standard \textsc{Select-Neighbors} rule is applied to determine whether the corresponding reciprocal edge should be added.
All upper layers, node levels, and the entry point remain unchanged.

\section{Extended Related Work}\label{app:related}
\noindent\textbf{Graph-based Approximate Nearest Neighbor Search.}
Approximate nearest neighbor search (ANNS) has been studied through several indexing paradigms, including partition- and quantization-based methods~\citep{10.1109/TPAMI.2010.57, 8733051} as well as graph-based methods~\citep{10.1109/TPAMI.2018.2889473, 10.14778/3303753.3303754, NEURIPS2019_09853c7f}, among which graph-based ANNS has emerged as one of the most effective and widely adopted approaches in practical tasks such document retrieval~\citep{ARSLAN20243781}, offering a strong balance between search quality and efficiency.
These methods typically construct a proximity graph over high-dimensional vector representations and answer a query by greedily traversing the graph toward its nearest neighbors~\citep{NEURIPS2019_09853c7f}.

Early graph-based approaches, such as FANNG~\citep{7780985} and EFANNA~\citep{fu2016efannaextremelyfast}, accelerate search through neighbor expansion over approximate KNN graphs. 
HNSW later introduced a hierarchical navigable small-world graph that enables highly effective greedy search and has become a standard index for high-recall in-memory vector retrieval~\citep{10.1109/TPAMI.2018.2889473}. 
NSG further improves this line of work by explicitly optimizing graph connectivity, path length, and index size, yielding a sparse yet highly navigable graph structure~\citep{10.14778/3303753.3303754}. 
At larger scales, DiskANN and its Vamana graph construction algorithm~\citep{NEURIPS2019_09853c7f} extend graph-based ANNS to billion-scale, SSD-resident settings, demonstrating strong empirical performance in large-scale retrieval despite limited theoretical guarantees~\citep{NEURIPS2023_d0ac28b7}. 
Building on this, FreshDiskANN supports real-time insertions and deletions~\citep{singh2021freshdiskannfastaccurategraphbased}, while Filtered-DiskANN further enables native support for filtered ANN queries~\citep{10.1145/3543507.3583552}.
Despite this progress, graph construction in existing ANNS methods remains fundamentally geometric and structural: edges are selected or pruned primarily based on vector proximity, graph sparsity, and navigability. 
As a result, these methods do not explicitly account for whether neighboring nodes are semantically redundant, complementary, or equally useful for downstream retrieval. 
Although stronger embedding models can improve the alignment between geometry and semantics~\citep{bge-m3, SFR-embedding-2, text-embedding-3-large}, a gap still remains between geometric proximity and semantic informativeness for retrieval~\citep{taghavi2026arguseyesassessingretrieval}. 
Our work complements this by introducing an \emph{offline} LLM-guided pruning paradigm on top of an existing vector-based ANN graph, where, instead of modifying query-time search, our method actively prunes local graph structure toward more semantically coherent neighborhoods while preserving the original navigation efficiency at test time.

\noindent\textbf{LLM-Augmented Retrieval at Test Time.}
A separate line of work improves document retrieval by introducing stronger reasoning at inference time, especially for complex queries where embedding similarity is insufficient for realistic tasks that require nontrivial reasoning to identify the most relevant documents~\citep{su2025bright}.
Existing LLM-augmented retrieval methods mainly improve retrieval quality in three directions. 
Some methods enhance the given queries by rewriting or reformulating the input before retrieval, with the goal of better aligning the search request with relevant documents~\citep{ma-etal-2023-query, chan2024rqrag}. 
Another line of work strengthens the retriever itself by learning reasoning-aware retrieval representations. Representative examples include INF-X-Retriever~\citep{inf-x-retriever-2025}, a single-stage dense retrieval framework for reasoning-intensive search that emphasizes intent-aligned retrieval, ReasonIR~\citep{shao2025reasonir}, which trains retrievers specifically for general reasoning tasks, and RaDeR~\citep{das-etal-2025-rader}, which develops reasoning-aware dense retrieval models using synthesized supervision from reasoning trajectories. 
Furthermore, some approaches improve retrieval by introducing stronger reasoning at the reranking stage, where direct rerankers rescore an initial candidate set returned by a first-stage retriever~\citep{sun2026divermultistageapproachreasoningintensive, zhang-etal-2025-rearank, liu2025reasonrankempoweringpassageranking}. 
For example, REARANK~\citep{zhang-etal-2025-rearank} performs listwise reasoning-based reranking over retrieved candidates.

While they show strong empirical performance, these approaches primarily inject additional semantic reasoning at test time, which increases latency and serving cost while leaving the fundamental mismatch between geometry and semantics in the underlying graph unresolved. 
In contrast, our method uses LLM reasoning offline to refine local graph connectivity and address this structural misalignment at its root, thereby improving the retrieval index itself and reducing the need for further LLM assistance at test time.

\noindent\textbf{Query-Time Graph Search Algorithms.}
Recent work has also explored improving graph-index retrieval from the query-time side by changing the search algorithm rather than the graph construction process. 
For example, \cite{xu2024bimetricframeworkfastsimilarity} builds the data structure using a cheap proxy metric but modifies the query procedure to approximate an expensive ground-truth metric, while reranker-guided search~\cite{xu2026beyond} alters query-time graph traversal to decide which candidates should be evaluated by a stronger reranker under a limited budget.
These methods are complementary to ours: they keep the graph largely unchanged but introduce a more involved online search procedure. 
In contrast, our method uses LLM reasoning offline to refine the graph structure itself, without requiring any change to the query-time search algorithm. 
As a result, our refined graph remains compatible with standard greedy search and can also be combined with reranker-guided or bi-metric query-time search strategies.

\end{document}